\documentclass[10pt,twocolumn,letterpaper]{article}

\usepackage{wacv}

\usepackage{comment}     
\usepackage{times}
\usepackage{latexsym}
\usepackage[T1]{fontenc}
\usepackage[utf8]{inputenc}
\usepackage{microtype}
\usepackage{inconsolata}
\usepackage{amsmath,amsfonts,amssymb}
\usepackage{graphicx}
\usepackage{xcolor}
\usepackage{bm}
\usepackage{algorithm}
\usepackage{algorithmic}
\usepackage{multirow}
\usepackage{booktabs}
\usepackage{tabularx}
\usepackage{adjustbox}
\usepackage{subcaption}
\usepackage{stfloats}
\usepackage{placeins}
\usepackage{xspace}

\definecolor{wacvblue}{rgb}{0.21,0.49,0.74}
\usepackage[pagebackref,breaklinks,colorlinks,allcolors=wacvblue]{hyperref}

\def\wacvPaperID{1914}
\def\confName{WACV}
\def\confYear{2027}

\usepackage[most]{tcolorbox}
\newtcblisting{promptbox}[1]{
  breakable,
  enhanced,
  colback=gray!5,
  colframe=black!55,
  boxrule=0.4pt,
  arc=2pt,
  title=#1,
  fonttitle=\bfseries\small,
  coltitle=white,
  colbacktitle=black!70,
  left=4pt,right=4pt,top=4pt,bottom=4pt,
  listing only,
  listing options={
    basicstyle=\ttfamily\footnotesize,
    breaklines=true,
    breakatwhitespace=false,
    breakindent=0pt,
    columns=fullflexible,
    keepspaces=true,
    showstringspaces=false,
    extendedchars=true,
    inputencoding=utf8,
    aboveskip=0pt,belowskip=0pt,
  }
}

\newcommand{\method}{{\fontfamily{lmtt}\selectfont FOLIO}\xspace}

\DeclareMathOperator*{\argmax}{arg\,max}
\newcommand{\clip}{\operatorname{clip}}

\begin{document}

\title{\method: Focused Semantic Memory for Streaming Video Understanding}

\author{Haoyang Fan$^{1}$ \quad
Dhruv Parikh$^{1}$ \quad
Anvitha Ramachandran$^{1}$ \quad
Sameh Gobriel$^{2}$\\
Nilesh Jain$^{2}$ \quad
Rajgopal Kannan$^{3}$ \quad
Viktor Prasanna$^{1}$\\
$^{1}$University of Southern California (USC), Los Angeles, CA, USA\\
$^{2}$Intel Labs, USA\\
$^{3}$DEVCOM Army Research Office\\
{\tt\small \{haoyangf,dhruvash,alramach,prasanna\}@usc.edu}\\
{\tt\small \{sameh.gobriel,nilesh.jain\}@intel.com}\\
{\tt\small rajgopal.kannan.civ@army.mil}
}

\maketitle

\begin{abstract}
In online streaming video understanding, a video stream continues to arrive and
queries may be issued at any time.
Because streaming frames grow without bound, the system must continuously
compress and retain information from the observed video prefix while future
frames and future queries remain unknown.
The core challenge is deciding what information to retain and how to organize
the maintained history: as this history grows with the stream, memory cost
increases and many redundant visual details are retained, whereas later queries
often depend on specific entities, actions, and their temporal changes.
To address this challenge, we introduce \method, a training-free focused
semantic memory system that assigns higher-detail records to selected observed
entities and actions while keeping surrounding context compact.
As the stream arrives, \method updates memory at the segment level, guided by a
dynamic focus state, combining a short-term visual buffer with a long-term
semantic memory organized around observed entities and linked to a
visual-evidence cache.
At query time, hybrid retrieval combines deterministic direct matching with
conditional VLM-assisted semantic query expansion.
With Qwen3-VL-8B, \method reaches 82.0\% and 69.1\% accuracy on the
evaluated Perception and Backward modes of OVO-Bench, respectively, and
74.5\% overall accuracy on StreamingBench. It substantially reduces
memory-writing cost by reserving detailed records for focused entities and
storing surrounding context compactly.
\end{abstract}

\section{Introduction}\label{sec:intro}
\begin{figure*}[t!]
\centering
\includegraphics[width=\textwidth]{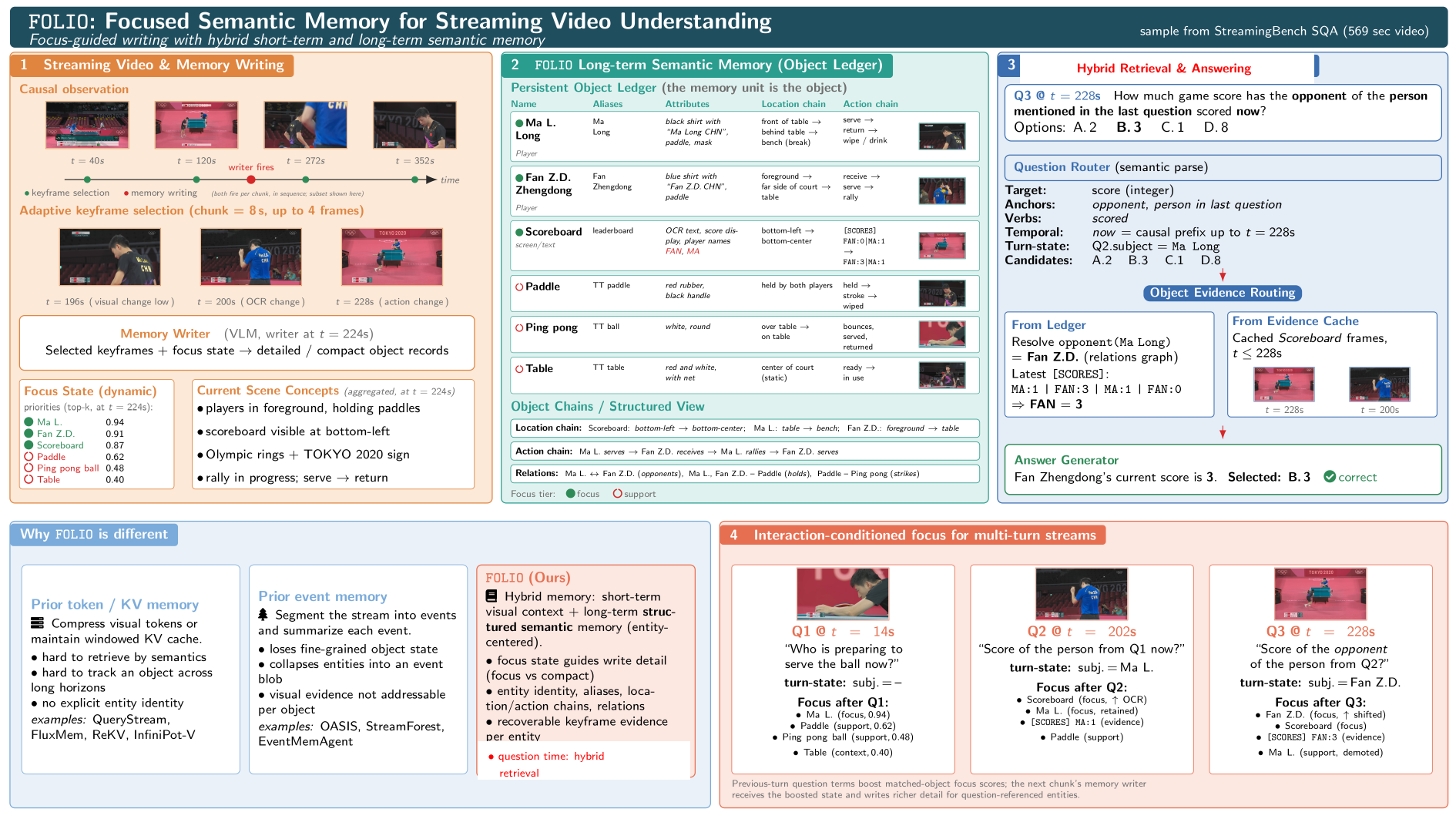}
\caption{\textbf{\method overview.} A training-free focused semantic memory system
for streaming video understanding.
\textbf{(1)}~Online segment-by-segment keyframe selection provides visual
evidence for focus-guided memory writing.
\textbf{(2)}~A hybrid memory combines a short-term visual buffer with a long-term
semantic memory linked to a visual-evidence cache.
\textbf{(3)}~At query time, \method uses hybrid retrieval to
link the query to relevant memory and evidence.
\textbf{(4)}~In multi-turn settings, previous-turn queries update the focus
state for subsequent memory writes.}
\label{fig:teaser}
\end{figure*}

Video-LLMs have made rapid progress on offline video understanding through
stronger backbones, video instruction tuning, and long-context
modeling~\cite{videollava,llavavideo,qwen2vl,qwen25vl,internvideo2,longva,
longvu,mvbench,egoschema,videomme,mlvu}, but offline evaluation assumes the
model can inspect the complete video before answering.
Streaming video understanding instead exposes an online setting in which a
video stream arrives over time and users may issue single- or multi-turn
queries at any moment.
This setting arises in applications such as autonomous
driving~\cite{drivelm}, robotic assistance~\cite{rt2}, and AR/wearable
agents~\cite{ego4d}, where a system must interpret the scene while it is still
unfolding.
At query time, the model can only answer from the observed prefix, while
future frames and future queries remain unavailable.
Because streaming frames grow without bound, the system must continuously
compress and retain information from the observed video prefix rather than
relying on full-video inspection.
Recent benchmarks~\cite{ovobench,streamingbench,svbench,ovbench,rtvbench}
make this constraint explicit through real-time perception, backward tracing,
and contextual multi-turn interaction, where a model must answer about
information that may no longer be visible.

Recent work addresses this challenge by improving how visual history is
retained, compressed, or scheduled: recent-window baselines~\cite{simplestream},
token/cache compression~\cite{querystream,fluxmem,freshmem,curvestream,
timechatonline,rekv,streamkv,infinipotv,stc}, scene/event/hierarchical
memories~\cite{oasis2026,streamforest,eventmemagent,vista,videoem}, and
proactive dialogue systems~\cite{videollmonline,mmduet,streambridge,livestar,
eyeswideopen,lionfs,streammind,vispeak,proassist}.
These methods avoid retaining all incoming frames by compressing or scheduling
visual inputs, but their memory units are typically frames, tokens, caches,
windows, scenes, or events.
First, the maintained history continues to grow with the stream, so memory
writing and later retrieval can become costly even after raw frames are
compressed.
Second, later queries are often sparse and target-specific, depending on
particular entities, actions, events, and their changes over time rather than
the entire stream.
Third, retained history is often organized around frames, windows, or events,
which can scatter target-related information across time and force query-time
retrieval to reconstruct the relevant temporal chain, adding latency and
retrieval noise.
Multi-turn interaction further amplifies these issues, because later queries
are asked over a longer horizon and may return to targets or events introduced
several turns earlier.

To address these issues, we introduce \method, a training-free focused semantic
memory system for streaming video understanding
(Fig.~\ref{fig:teaser}).
The key idea is focused memory construction: selected observed entities and
actions receive richer records, while surrounding context is kept compact.
As the stream arrives, \method updates a hybrid memory at the segment level,
guided by a dynamic focus state.
The hybrid memory combines a short-term visual buffer, which keeps recent
observations available, with a long-term semantic memory implemented as an
entity-centered structured memory.
The long-term memory stores target states, locations, relations, and associated
events over time, and links these semantic records to a visual-evidence cache
so that relevant frames can be recovered when needed.
At query time, \method uses hybrid retrieval, combining deterministic direct
matching with conditional VLM-assisted semantic query expansion.
Since multi-turn queries often revolve around the same entities or actions,
previous turns update the focus state so that later memory construction can
continue tracking targets under discussion.
\textcolor{black}{\method improves both accuracy and memory-writing efficiency on
the evaluated Perception and Backward modes of OVO-Bench and on StreamingBench.
}
Our contributions are:
\begin{itemize}
    \item We formulate online memory construction for streaming video
    understanding as deciding what information to retain and how to organize
    the maintained history before future queries are known.
    \item We introduce a training-free focused semantic memory system whose
    dynamic focus state prioritizes entities and actions for richer memory
    updates, keeps surrounding context compact, and raises the priority of
    targets mentioned in previous turns for multi-turn query streams.
    \item We design a hybrid memory and retrieval pipeline that combines a
    short-term visual buffer, a long-term semantic memory implemented as an
    entity-centered structured memory, a
    visual-evidence cache, and hybrid retrieval with conditional VLM-assisted
    semantic query expansion.
    \item We validate \method on the evaluated Perception and Backward modes of
    OVO-Bench and on StreamingBench, showing improved accuracy while
    substantially reducing the cost of maintaining streaming memory.
\end{itemize}

\section{Related Work}\label{sec:related}

\noindent\textbf{Video-LLMs and long-video understanding.}
Video-LLMs~\cite{videollava,llavavideo,qwen2vl,qwen25vl,qwen3vl,
internvideo2,videollama,videochat,mvbench,egoschema,videomme,mlvu,lvbench,
nextqa} and long-video systems~\cite{longva,llamavid,longvu,videoxl,
videoxltwo,moviechat,malmm,videostreaming,videoagent,videotree,goldfish,
videorag} typically assume access to the full video or query before
selecting evidence; \method writes memory online from the observed prefix.

\noindent\textbf{Streaming video understanding and interaction.}
Streaming benchmarks~\cite{ovobench,streamingbench,svbench,ovbench,rtvbench,
livestar,eyeswideopen} and online/proactive/streaming-reasoning
systems~\cite{videollmonline,videollmmod,streamchat,streamingvlm,
streambridge,aura,livecc,internlmomnilive,mmduet,livestar,
lionfs,streammind,vispeak,egospeak,proassist,omnimmi,
thinkwhilewatching2026,videostreamingthinking2026,
thinkingstreamingvideo2026,thinkasyousee2026,wat2026,speakwhilewatching2026}
study online observation, when to speak, and streaming reasoning.
These systems ask how to process streams efficiently, how to align training
with streaming inference, or when a model should respond.
\method is complementary: it asks what information should be retained, how it
should be organized, and how interaction should affect later memory
construction.

\noindent\textbf{Memory, retrieval, and compression for streaming video.}
Prior work retains history through visual-token, feature, or KV-cache
compression~\cite{querystream,fluxmem,freshmem,curvestream,timechatonline,
stc,rekv,streamkv,infinipotv,streammem,livevlm,streamingtom,hermes} or recent
frame windows~\cite{simplestream}.
Another family organizes history into temporal, scene, event, or hierarchical
memories: OASIS maintains short and medium visual windows plus an on-demand
hierarchical event forest; StreamForest builds a persistent event-memory
forest; EventMemAgent adds adaptive tool use over event-centric memory; and
Vista, Event-VStream, and related systems use scene or event units as the
online memory abstraction~\cite{oasis2026,streamforest,eventmemagent,vista,
eventvstream,hierarchicaleventmemory,videoem}.

\noindent\textbf{Our difference.}
\textcolor{black}{Entity-, graph-, and event-centered memories have already been
explored in prior video-understanding systems. ESOM stores online object tracks
for visual-query localization, GraphVideoAgent constructs entity-relation
graphs for query-guided frame selection, and Semantic Event Graphs represent
human--object interactions as temporal event logs~\cite{esom,graphvideoagent,
semanticeventgraphs}. Complementary work studies visually aligned retrieval,
multi-source multimodal memory, and adaptive visual-memory compression, as in
Video-RAG, WorldMM, and FreshMem~\cite{videorag,worldmm,freshmem}.
\method's entity-centered memory is in the spirit of the above approaches. However, a core distinction of \method is its online write-time detail allocation before future queries are known. Our proposed dynamic focus state determines which observed entities and actions receive richer predefined records. Entity-centered long-term semantic memory
provides the organizational substrate for these updates, while previous turns
can alter the focus state used for later stream segments. At query time, hybrid
retrieval combines deterministic direct matching with conditional VLM-assisted
semantic query expansion over this structured memory}. See Appendix~\ref{app:extended-related-work} for an extended related work discussion.

\section{Method}\label{sec:method}
\begin{figure*}[t!]
\centering
\includegraphics[width=\textwidth]{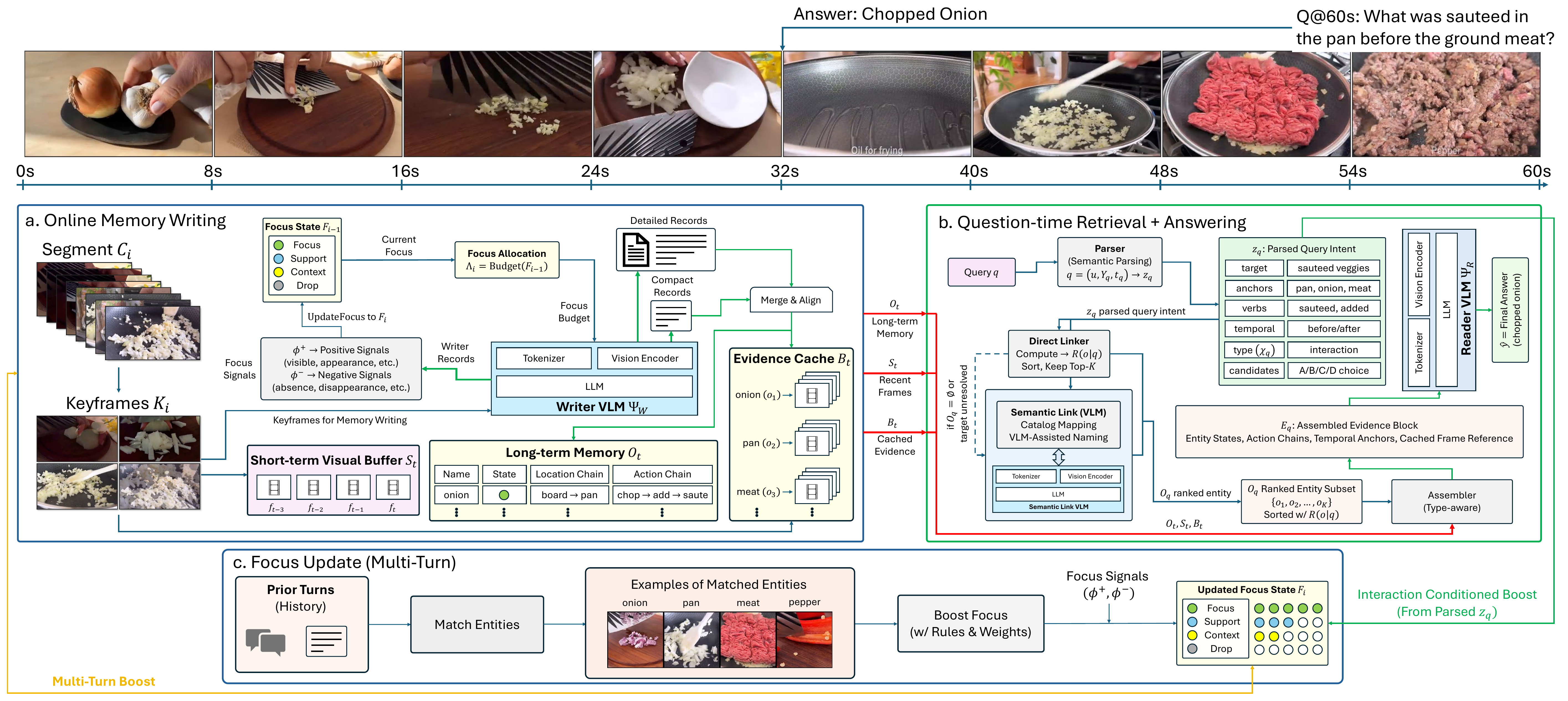}
\caption{\textbf{\method overview.}
A streaming cooking video runs across the top with an illustrative query at
$t_q{=}60$\,s.
\textbf{(a)} For each segment $C_i$, a writer VLM produces detailed and compact
records under focus-guided writing levels
$\Lambda_i{=}\textsc{Budget}(\mathcal{F}_{i-1})$ and updates the hybrid memory
$\mathcal{M}_i{=}(\mathcal{S}_i,\mathcal{O}_i,\mathcal{B}_i)$;
$\textsc{UpdateFocus}$ refreshes $\mathcal{F}_i$ from observation and
interaction signals.
\textbf{(b)} A query is linked to a ranked entity subset $\mathcal{O}_q$, using
VLM-assisted semantic query expansion when direct matching is insufficient,
and assembled into evidence $\mathcal{E}_q$ for an answering VLM.
\textbf{(c)} Across turns, entities and actions matched from prior queries
contribute interaction-relevance signals to future focus updates.}
\label{fig:method-overview}
\end{figure*}

\subsection{Problem Setup and Online Memory Construction}
\label{sec:problem-overview}

Online streaming video understanding considers a video stream
$V=\{f_t\}_{t\geq 1}$ that unfolds over time, with queries arriving before the
stream necessarily ends.
At query time $t_r$, the system may use only the observed prefix
$V_{\leq t_r}$ and, in multi-turn settings, the preceding dialogue history
$\mathcal{H}_{r-1}$.
Future frames and future queries are unavailable.

\method processes the observed stream as ordered, fixed-duration segments
$C_1,C_2,\ldots$.
A segment is processed once it has completely arrived; segmenting therefore
specifies the online update granularity rather than granting access to the
complete video.
After segment $C_i$, \method maintains memory $\mathcal{M}_i$ and focus state
$\mathcal{F}_i$.
Let $\tau_r$ denote the index of the last completed segment before query time
$t_r$; the corresponding long-term memory available to the query is therefore
$\mathcal{O}_{\tau_r}$.

As shown in Fig.~\ref{fig:method-overview}, the previous focus state
$\mathcal{F}_{i-1}$ determines a writing budget for segment $C_i$.
A keyframe selector first chooses a compact visual set $\mathcal{K}_i$.
The writer VLM is then invoked once for the completed segment and generates
independent detailed and compact records from $\mathcal{K}_i$.
The memory system, rather than the VLM, assigns writing levels, matches
identities, merges records into persistent entity chains, and updates
$\mathcal{F}_i$ for the next segment.
At query time, \method combines recent frames, retrieved structured records,
and conditionally recovered historical keyframes.
The complete online construction procedure is provided in
Appendix~\ref{app:method-details}.

\subsection{Focus State for Memory Writing}
\label{sec:focus-writing}

The focus state is a deterministic, training-free state derived from recent
observation and activity signals for tracked entities.
It controls only the description budget used for later memory updates; it does
not predict future questions, and the writer VLM does not decide which entities
receive detailed records.
After the records for $C_i$ have been merged, \method updates the score of each
tracked entity $o$ by decaying its previous score and adding evidence from the
current segment:
\begin{equation}
p_i(o)=\operatorname{clip}_{[0,1]}\left(
\gamma p_{i-1}(o)
+\boldsymbol{\alpha}^{\top}\boldsymbol{\phi}^{+}_i(o)
-\boldsymbol{\beta}^{\top}\boldsymbol{\phi}^{-}_i(o)
\right).
\end{equation}
The positive signals $\boldsymbol{\phi}^{+}_i(o)$ include first appearance,
reappearance, persistence, location or state change, event participation,
manipulable-object status, and relevance to previous interactions.
The negative signals $\boldsymbol{\phi}^{-}_i(o)$ capture absence, prolonged
disappearance, and static background behavior.
\textcolor{black}{The decay, signal weights, writing thresholds, and writing
budgets are fixed across datasets rather than learned relevance scores. Their
exact values are reported in Appendix~\ref{app:fixed-parameters}.}
The memory system ranks retained entities by this score and maps them to
focus, support, or context groups for writing.
Entities in the focus group receive detailed records, while support entities and persistent
scene context receive compact records.
Additional feature and writing-level details are provided in
Appendix~\ref{app:focus-details}.
These groups allocate detail but do not close object discovery.
The writer must still report newly visible manipulated objects, tools,
containers, appliances, text regions, and unusual objects.
Consequently, focus affects the richness of future updates for tracked
entities without preventing a new query target from entering memory on first
appearance.

\subsection{Hybrid Semantic Memory Structure}
\label{sec:hybrid-memory}

\method maintains
\begin{equation}
\mathcal{M}_i=(\mathcal{S}_i,\mathcal{O}_i,\mathcal{B}_i),
\end{equation}
where the three components preserve complementary evidence.
The short-term visual buffer $\mathcal{S}_i$ retains four recent frames near
the current stream time.
It preserves visual detail for current-state questions that may be lost in
textual records.
The long-term semantic memory $\mathcal{O}_i$ is implemented as an
entity-centered structured memory composed of persistent entity slots.
Each slot contains a stable identifier, canonical name, aliases, category,
attributes, and time-ordered observations of state, location, holder,
relations, visible text, and associated actions or events.
The slot therefore exposes a retrievable temporal trace rather than a flat
collection of segment summaries.
The visual-evidence cache $\mathcal{B}_i$ stores selected historical keyframes
with timestamps and links them to the corresponding entity and event records.
It is not concatenated unconditionally at query time; it supplies visual
grounding only when the retrieval policy requests historical evidence.

\subsection{Focus-Guided Online Memory Writing}
\label{sec:online-writing}

For each completed segment, \method selects boundary frames and additional
high-change evidence when needed.
The exact keyframe rule is given in Appendix~\ref{app:keyframe-chain-details}.
The selected keyframes and the writing budget are passed to one writer-VLM
invocation.
Detailed records describe state and location changes, holder, relations,
visible text, interactions, and supporting frame indices.
Compact records retain the entity name, category, and a short
state/location/relation note.
This shared call lets the writer preserve focused entities in greater detail
while retaining lightweight context.
The writer first emits independent per-segment records.
A deterministic post-write matcher then aligns each record to the existing
entity slots.
Names are normalized by removing punctuation and generic modifiers and by
applying a fixed alias dictionary.
Exact canonical-name and alias matches receive confidence 1.0 and 0.95.
A shared head noun with the same category receives 0.8; a head-noun match
without category evidence receives 0.6; and normalized-name containment
receives 0.5.
The match is accepted at confidence at least 0.5.
For weaker head-noun or partial matches, conflicting color attributes reject
the merge.
A matched record is appended to the existing entity's temporal trace; an
unmatched record starts a new slot.
This operation organizes observations produced independently across segments
and reduces serialized redundancy.
It is not an additional VLM call and does not claim that merging itself
improves answer accuracy.
The selected keyframes and record-to-frame links are appended to the
visual-evidence cache in the same update.

\subsection{Hybrid Retrieval and Answering}
\label{sec:evidence-routing}

When query $q_r$ arrives, \method first parses entity names, action terms,
temporal anchors, and answer-option keywords.
Deterministic direct matching scores these fields against canonical names,
aliases, categories, attributes, and action chains in
$\mathcal{O}_{\tau_r}$, and retains at most five entity records.
This direct path does not invoke a VLM.
VLM-assisted semantic query expansion is triggered only when direct matching
returns no entity.
It makes one conditional VLM call using a compact catalog of the identifiers,
names, aliases, categories, representative locations, and actions already
stored in memory.
This call is text-only and performs a narrow catalog-linking task: it returns a
bounded set of existing identifiers with a short rationale and option prior,
without inspecting video frames or producing the final answer.
Semantic expansion may return up to five existing entity or event identifiers;
it cannot introduce a new memory item, generate an answer, or add evidence not
present in the catalog.
The returned identifiers are resolved through the same deterministic evidence
assembler used by direct matching.
Thus, semantic query expansion increases recall for generic or indirect
references while preserving the boundary between retrieval and answering.
\method next assembles the selected structured records with four recent
frames.
Historical keyframes are recovered only when the configured grounding checks
indicate that textual retrieval is empty or insufficient.
Recovery considers at most three retrieved entities and adds at most four
cached frames.
The answering VLM is then invoked once with the question, options, assembled
records, recent frames, and any recovered frames.
Query-time calls use the 24K maximum model context configured in the
evaluation, and answer generation is capped at 1,024 output tokens.
We report direct-matching, semantic-expansion, cache-recovery, and answering
costs separately because semantic expansion and answering are distinct VLM
invocations.

\subsection{Focus State Updates for Multi-Turn Queries}
\label{sec:multiturn-extension}

In a multi-turn stream, each query is answered from the memory available at
its arrival time.
After answering, \method links entity and action mentions from that completed
turn to the current entity-centered memory.
\textcolor{black}{Matched entities receive a fixed interaction-relevance boost to
their focus score, which raises their priority for higher-detail writing in
later completed segments.}
Unmatched mentions are retained as pending references and reconsidered when
new entities appear.
This update changes only the writing budget for future completed segments: it
neither rewrites past records nor accesses future frames or later queries.
If the segment currently arriving has not finished writing, its long-term
records are unavailable to that answer, while its recent frames may still be
present in the short-term visual buffer.

\section{Experiments}\label{sec:exp}

\begin{table*}[t]
\centering
\caption{Single-turn query stream results on the full Perception and Backward
subsets of OVO-Bench. Baseline task scores are quoted from the corresponding
papers. Avg is recomputed as the unweighted mean of the displayed task scores
within each mode for all methods.}
\label{tab:ovo_main}
\footnotesize
\setlength{\tabcolsep}{2.2pt}
\renewcommand{\arraystretch}{0.92}
\resizebox{0.79\textwidth}{!}{%
\begin{tabular}{l|ccccccc|cccc}
\toprule
\multirow{2}{*}{\textbf{Model}} & \multicolumn{7}{c|}{\textbf{Perception}} & \multicolumn{4}{c}{\textbf{Backward}} \\
\cmidrule(lr){2-8} \cmidrule(l){9-12}
 & \textbf{OCR} & \textbf{ACR} & \textbf{ATR} & \textbf{STU} & \textbf{FPD} & \textbf{OJR} & \textbf{Avg} & \textbf{EPM} & \textbf{ASI} & \textbf{HLD} & \textbf{Avg} \\
\midrule
Gemini 1.5 Pro & 85.9 & 67.0 & 79.3 & 58.4 & 63.4 & 62.0 & 69.3 & 58.6 & 76.4 & 52.6 & 62.5 \\
GPT-4o & 69.8 & 64.2 & 71.6 & 51.1 & 70.3 & 59.8 & 64.5 & 57.9 & 75.7 & 48.7 & 60.8 \\
LLaVA-Video-7B & 69.1 & 58.7 & 68.8 & 49.4 & 74.3 & 59.8 & 63.4 & 56.2 & 53.5 & 51.5 & 53.7 \\
LLaVA-OneVision-7B & 66.4 & 57.8 & 73.3 & 53.4 & 71.3 & 62.0 & 64.0 & 56.2 & 55.4 & 21.5 & 44.4 \\
Qwen2-VL-7B & 40.4 & 50.5 & 63.8 & 47.2 & 66.3 & 53.6 & 53.6 & 48.7 & 35.5 & 45.0 & 43.1 \\
Qwen2-VL-72B & 65.8 & 60.6 & 69.8 & 51.7 & 69.3 & 54.4 & 61.9 & 52.5 & 60.8 & 57.5 & 56.9 \\
Flash-VStream-7B & 24.2 & 29.4 & 28.5 & 33.7 & 25.7 & 28.8 & 28.4 & 39.1 & 37.2 & 5.9 & 27.4 \\
VideoLLM-online-8B & 8.1 & 2.8 & 12.0 & 14.0 & 45.5 & 21.1 & 17.3 & 18.8 & 12.2 & 17.7 & 16.2 \\
StreamChat & 51.7 & 47.7 & 58.6 & 41.0 & 61.1 & 47.8 & 51.3 & 50.2 & 54.1 & 37.6 & 47.3 \\
Dispider & 57.7 & 49.5 & 62.1 & 44.9 & 61.4 & 51.6 & 54.5 & 48.5 & 55.4 & 45.4 & 49.8 \\
Qwen2.5-VL-7B
& 76.5 & 57.8 & 68.1 & 46.6 & 66.3 & 56.5 & 62.0 & 49.8 & 58.1 & 44.6 & 50.8 \\
+ OASIS & 85.2 & 72.5 & 66.4 & 52.2 & 67.3 & 64.7 & 68.1 & 51.9 & 58.8 & 48.9 & 53.2 \\
Qwen3-VL-8B
& 83.9 & 58.7 & 70.7 & 56.2 & 69.3 & 64.1 & 67.2 & 51.2 & 60.8 & 43.6 & 51.9 \\
+ OASIS & 92.0 & 80.7 & 81.0 & 67.4 & 67.3 & 79.9 & 78.1 & 62.0 & 60.1 & 47.3 & 56.5 \\
\midrule
Qwen2.5-VL-7B
 + \method
& 95.3 & 83.3 & 87.5 & 60.9 & 62.9 & 72.3 & 77.0
& 50.5 & 54.3 & 66.7 & 57.2 \\
Qwen3-VL-8B
 + \method
& 96.7 & 85.7 & 85.7 & 70.6 & 62.5 & 90.9 & 82.0 & 59.0 & 63.6 & 84.6 & 69.1 \\
\bottomrule
\end{tabular}
}
\end{table*}

\subsection{Experimental Setup}
\label{sec:experimental-setup}

\noindent\textbf{Streaming protocol.}
\textcolor{black}{We evaluate \method{} under a strict online streaming protocol.
Videos are processed as ordered segment streams. At query time, \method{} uses
only the observed video prefix and memory written from completed segments up to
that time. Future frames and future queries are never used, and answer options
are provided only at query time. In multi-turn evaluation, previous model
predictions are used as context rather than ground-truth answers.}

\noindent\textbf{Benchmarks.}
\textcolor{black}{Our main experiments use OVO-Bench~\cite{ovobench} and
StreamingBench~\cite{streamingbench}. We evaluate the complete
\textit{Perception} and \textit{Backward} subsets of OVO-Bench for single-turn
online perception and retrospective recall. StreamingBench provides multi-turn
interaction, with multiple timestamped questions per video, allowing us to
evaluate whether memory and prior turns support later queries. We report the
Realtime and OmniSource aggregates together with SQA and Proactive Output. The
supplement additionally evaluates StreamBench~\cite{streambench}, an open-ended
multi-round benchmark spanning egocentric, web, and movie videos.}

\noindent\textbf{Models and baselines.}
We evaluate \method with Qwen2.5-VL-7B and Qwen3-VL-8B backbones. We compare
against OASIS~\cite{oasis2026} for single-turn streaming QA on OVO-Bench and
Think-While-Watching~\cite{thinkwhilewatching2026} for multi-turn streaming QA
on StreamingBench.
\textcolor{black}{Baseline task scores are quoted from prior work;
Table~\ref{tab:ovo_main} averages are recomputed from the displayed scores.
Latencies in Table~\ref{tab:matched_latency_main} are measured by us.}

\noindent\textbf{VLM setting.}
We serve all VLMs with vLLM~\cite{kwon2023efficient} using one GPU per model
instance through an
OpenAI-compatible multimodal interface. The same serving configuration is used
for all VLM calls in memory writing, semantic query expansion, and answer
generation, with a 24K-token maximum model length.

\noindent\textbf{Hardware.}
All experiments are conducted on NVIDIA RTX 6000 Ada Generation GPU servers,
with 48\,GB memory and 960\,GB/s memory bandwidth per GPU.


\subsection{Main Results}
\label{sec:main-results}

\paragraph{Single-turn results.}
Table~\ref{tab:ovo_main} reports per-task accuracy on the Perception and
Backward subsets of OVO-Bench. \method improves over OASIS on both
backbones, reaching $77.0/57.2$ Perception/Backward average with
Qwen2.5-VL-7B and $82.0/69.1$ with Qwen3-VL-8B. The gains are
especially clear on Backward queries, where the stronger backbone
widens the margin over OASIS from $+4.0$ to $+12.6$ points. The largest
improvement appears on hallucination detection (HLD), suggesting that
the structured memory helps distinguish whether an entity or event has
actually appeared in the observed stream. HLD also benefits from conservative
answer calibration, so we report it separately and do not use it as the sole
evidence for memory quality. The main exception is FPD,
where \method trails OASIS slightly; this task requires extrapolating
beyond the observed prefix and is less directly served by memory
retrieval.
\textcolor{black}{We do not evaluate Forward Active Responding (FAR) because it
uses a distinct forward-active protocol that repeatedly queries the model as
future frames arrive, rather than answering an explicit query from the observed
prefix.}

\paragraph{Multi-turn results.}
Table~\ref{tab:streamingbench_main} evaluates whether the memory design
\textcolor{black}{transfers to multi-turn streaming interaction on StreamingBench.} \method
improves over both the online backbone and the strongest multi-turn baseline,
Think-While-Watching. The
\textcolor{black}{largest gains appear on Realtime and OmniSource. These results
show stronger performance on queries over the evolving stream, while the
controlled ablations in Sec.~\ref{sec:long-term-role} separately examine the
contribution of long-term memory.}
SQA remains more challenging because many queries require previous-turn
reference resolution and fine-grained reading of on-screen states such as
scores. This suggests that interaction-state tracking and detailed state
capture remain important directions for improvement.
\textcolor{black}{Although query-driven QA is our primary setting, we include
Proactive Output as an auxiliary result showing that the same streaming memory
can also support response-timing decisions.}
\begin{table*}[t]
\centering
\caption{\textcolor{black}{Multi-turn query stream results on StreamingBench.} For \method,
we report results with Qwen2.5-VL-7B and Qwen3-VL-8B; columns report the
Realtime and OmniSource aggregate aspects, the SQA and Proactive Output
subtasks, and overall accuracy. \textcolor{black}{External-baseline accuracies are
quoted from the corresponding papers.}}
\label{tab:streamingbench_main}
\small
\setlength{\tabcolsep}{3.5pt}
\renewcommand{\arraystretch}{0.95}
\resizebox{0.85\textwidth}{!}{%
\begin{tabular}{l|ccccc}
\toprule
\textbf{Method} & \textbf{Realtime} & \textbf{OmniSource} & \textbf{SQA}
& \textbf{Proactive} & \textbf{Overall} \\
\midrule
VideoLLM-online-8B & 35.99 & 28.45 & 30.80 & 3.92 & 32.48 \\
Dispider-7B & 67.63 & 35.66 & 34.80 & 25.34 & 53.12 \\
StreamAgent-7B & 74.28 & 36.26 & 39.60 & 28.90 & 57.02 \\
Qwen3-VL-8B (online instruct) & 25.12 & 21.47 & 17.20 & 17.60 & 23.04 \\
Think While Watching (Qwen3-VL-8B, multi-turn) & 74.40 & 37.47 & 50.00 & 40.00 & 58.82 \\
\midrule
Qwen2.5-VL-7B + \method & 75.3 & 61.8 & 51.6 & 23.1 & 69.7 \\
Qwen3-VL-8B + \method & \textbf{81.2} & \textbf{63.6} & \textbf{54.8} & \textbf{51.6} & \textbf{74.5} \\
\bottomrule
\end{tabular}
}
\end{table*}

\subsection{\textcolor{black}{Focus-Guided Memory Writing}}

\textcolor{black}{We analyze memory-writing cost on an OVO-Bench evaluation
subset obtained by randomly sampling approximately 30\% of the benchmark while
preserving its original question-type distribution.} The
savings come from two system choices: focus-guided memory writing,
which avoids writing every observed entity with the same level of detail, and
adaptive keyframe selection, which reduces the visual input sent to the writer
VLM. Table~\ref{tab:memory_construction_cost_main} compares these choices
against fixed visual input and uniform memory writing. Relative to the
fixed-uniform variant, \method reduces writer input tokens by 33.2\%, writer
output tokens by 31.8\%, written records by 31.3\%, and merged memory size by
32.4\%, while improving accuracy by 4.4 points on this evaluation subset. This
table evaluates the writing-allocation policy rather than serving as a clean
ablation of every component associated with the focus state.

\begin{table}[!t]
\centering
\caption{Accuracy and memory-writing cost under fixed/adaptive keyframe
selection and uniform/focus-guided writing. Merged bank reports the serialized
memory size after segment-level records are merged.}
\label{tab:memory_construction_cost_main}
\scriptsize
\setlength{\tabcolsep}{2.0pt}
\renewcommand{\arraystretch}{0.95}
\resizebox{\columnwidth}{!}{%
\begin{tabular}{lccccc}
\toprule
\textbf{Variant} & \textbf{Acc.} & \textbf{Writer} & \textbf{Writer} &
\textbf{Records} & \textbf{Merged bank} \\
 & \textbf{(\%)} & \textbf{input tok./Q} & \textbf{output tok./Q} &
\textbf{/Q} & \textbf{KB/Q} \\
\midrule
Fixed + uniform & 62.7 & 201.5K & 38.4K & 259.7 & 226.8 \\
Adaptive + uniform & 63.2 & 126.7K & 38.3K & 259.6 & 226.5 \\
Fixed + focus-guided & 65.9 & 207.3K & 26.4K & 179.2 & 154.9 \\
\method & \textbf{67.1} & \textbf{134.7K} & \textbf{26.2K} & \textbf{178.3} & \textbf{153.4} \\
\bottomrule
\end{tabular}
}
\end{table}

{\color{black}
\paragraph{Focus-state robustness and target coverage.}
We evaluate parameter sensitivity on an approximately 30\% sample of
OVO-Bench Backward, randomly drawn while preserving the original question-subtype
distribution. All variants use the same selected frames, writer and reader
models, memory schema, and evaluation protocol, while memory is regenerated
under each Focus setting. Scaling the writing thresholds from $0.5\times$ to
$1.5\times$ changes accuracy by at most 1.7 percentage points and memory size by
less than 7\%. Jointly halving the interaction, movement, and state-change score
updates does not reduce accuracy on this diagnostic (65.7\% versus 65.0\% with
the default values).
These results support local robustness to the writing thresholds and major
activity-related updates, rather than independent insensitivity or optimality
of every coefficient. In a separate audit of EPM traces, where question targets
can be directly aligned to visible entities, the focus state assigns the target
to either detailed or compact writing at least once for 81.3\% of evaluable
questions, including 75.0\% with at least one detailed allocation. Per-question
coverage averages each question's visible-chunk hit rate equally and reaches
64.4\%. Pooling all target-visible chunks instead gives 88.9\%, weighting
frequently recurring targets more heavily.
}

\begin{table}[!t]
\centering
\caption{\textcolor{black}{Focus parameter sensitivity and target-selection
coverage. Detailed/compact reports the numbers of records per segment.}}
\label{tab:focus_analysis_main}
\color{black}
\footnotesize
\setlength{\tabcolsep}{3.0pt}
\renewcommand{\arraystretch}{1.0}
\begin{tabular*}{\columnwidth}{@{\extracolsep{\fill}}lccc@{}}
\toprule
\multicolumn{4}{c}{\textbf{(a) Parameter sensitivity}} \\
\textbf{Setting} & \textbf{Acc. (\%)} & \textbf{Detailed/compact} & \textbf{KB/video} \\
\midrule
Threshold $\times0.5$ & 65.0 & 2.40/1.40 & 96.77 \\
Default & 65.0 & 2.39/1.40 & 93.30 \\
Threshold $\times1.5$ & 65.0 & 2.38/1.49 & 98.07 \\
Activity updates $\times0.5$ & 65.7 & 2.39/1.44 & 98.52 \\
\bottomrule
\end{tabular*}
\vspace{2pt}

\begin{tabular*}{\columnwidth}{@{\extracolsep{\fill}}lr@{}}
\toprule
\multicolumn{2}{c}{\textbf{(b) Target-selection audit}} \\
\textbf{Metric} & \textbf{Result} \\
\midrule
Any allocation at least once & 81.3\% \\
Detailed allocation at least once & 75.0\% \\
Mean per-question visible-chunk coverage & 64.4\% \\
Pooled visible-chunk coverage & 88.9\% \\
\bottomrule
\end{tabular*}
\end{table}

\paragraph{Component ablations.}
Figure~\ref{fig:memory_ablation_drop} ablates the main focus-memory
components on the same StreamingBench evaluation subset; the full
numeric table is in Table~\ref{tab:memory_ablation_app} in the appendix.
Removing the recent
visual window causes the largest drop, showing that short-term visual evidence
is indispensable for boundary and current-state queries. This is an important
boundary of the method: \method is not designed to replace recent visual
grounding, but to make earlier entities and their histories retrievable. Since
the sampled split contains many real-time queries, components that mainly help
longer-horizon memory can have smaller aggregate effects. Removing structured
memory while retaining chunk-level event/text retrieval is therefore less
harmful on this mixed split because many queries can still be answered from
local or event-level evidence. Removing the focus state or adaptive visual
writing reduces accuracy, supporting the role of salience-aware memory writing
and visual evidence selection. Section~\ref{sec:long-term-role} separately
examines long-term-memory behavior and entity matching.

\begin{figure}[t]
\centering
\scriptsize
\resizebox{0.92\columnwidth}{!}{%
\begin{tikzpicture}[x=0.102cm,y=0.34cm]
    \foreach \x/\lab in {0/30,10/40,20/50,30/60,40/70,50/80} {
        \draw[gray!18] (\x,0.45) -- (\x,6.65);
        \draw[gray!45] (\x,0.45) -- (\x,0.30);
        \node[anchor=north,gray!70] at (\x,0.18) {\lab};
    }
    \draw[densely dashed,gray!65] (41.0,0.45) -- (41.0,6.65);
    \node[anchor=north,gray!70] at (25,-0.45) {Accuracy (\%)};

    \node[anchor=east] at (-1.1,6) {w/o Recent Window};
    \fill[blue!60!black] (0,5.72) rectangle (2.4,6.28);
    \node[anchor=west] at (3.2,6) {\textbf{32.4}};
    \node[anchor=west,gray!70] at (10.2,6) {$-38.6$};

    \node[anchor=east] at (-1.1,5) {w/o Focus State};
    \fill[cyan!55!blue] (0,4.72) rectangle (35.5,5.28);
    \node[anchor=east,text=white] at (34.3,5) {\textbf{65.5}};
    \node[anchor=west,gray!70] at (36.7,5) {$-5.5$};

    \node[anchor=east] at (-1.1,4) {w/o Struct. Memory};
    \fill[teal!65!black] (0,3.72) rectangle (38.3,4.28);
    \node[anchor=east,text=white] at (37.1,4) {\textbf{68.3}};
    \node[anchor=west,gray!70] at (39.5,4) {$-2.7$};

    \node[anchor=east] at (-1.1,3) {w/o Adaptive Visual};
    \fill[green!55!black] (0,2.72) rectangle (38.3,3.28);
    \node[anchor=east,text=white] at (37.1,3) {\textbf{68.3}};
    \node[anchor=west,gray!70] at (39.5,3) {$-2.7$};

    \node[anchor=east] at (-1.1,2) {w/o Interaction Focus};
    \fill[orange!75!black] (0,1.72) rectangle (39.7,2.28);
    \node[anchor=east,text=white] at (38.5,2) {\textbf{69.7}};
    \node[anchor=west,gray!70] at (40.9,2) {$-1.3$};

    \node[anchor=east] at (-1.1,1) {Full};
    \fill[black!70] (0,0.72) rectangle (41.0,1.28);
    \node[anchor=east,text=white] at (39.8,1) {\textbf{71.0}};

    \draw[gray!55] (0,0.45) -- (50,0.45);
\end{tikzpicture}
}
\caption{Component ablation using Qwen3-VL-8B on StreamingBench. Bars show
accuracy after removing each component; gray
numbers report the drop relative to the full system.}
\label{fig:memory_ablation_drop}
\end{figure}

{\color{black}
\paragraph{Interaction-conditioned writing.}
On the complete StreamingBench SQA split, interaction-conditioned focus raises
accuracy from 53.6 to 54.8 when prior turns are included in the answer prompt,
and from 53.2 to 54.0 when they are removed. Its modest gain persists without
direct dialogue access, so we treat it as an auxiliary extension rather than a
standalone contribution. Table~\ref{tab:sqa_interaction_focus_app}
(Appendix~\ref{app:sqa-diagnostic}) gives the full $3\times2$ comparison.
}

\subsection{\textcolor{black}{Role and Reliability of Long-Term Memory}}
\label{sec:long-term-role}

{\color{black}
\paragraph{Beyond recent frames.}
We evaluate complementary views of long-term-memory behavior on OVO-Bench
Backward Tracing. Overall accuracy follows the type-proportional half split
used in our original submission, while the HLD column evaluates all HLD
questions in the complete valid Backward set. HLD accounts for 28.3\% of that
set and directly tests history verification, for which evidence may no longer
be available in recent frames. Calls per query and input tokens are averaged
over the complete valid Backward set. As shown in
Table~\ref{tab:long_term_role_main}, direct long-term retrieval raises HLD
accuracy from 51.8 to 75.0, and the full query policy reaches 80.4. On the
type-proportional overall split, the full policy reaches 56.7 compared with
54.5 using recent frames. The full query policy conditionally routes and
assembles evidence; it is a system-policy comparison rather than a fourth
single-variable evidence-source ablation because it also changes the answer
prompt and retry routing.
}

\begin{table}[H]
\centering
\caption{\textcolor{black}{Role of long-term semantic memory beyond the
short-term visual buffer on OVO-Bench Backward Tracing.}}
\label{tab:long_term_role_main}
\color{black}
\footnotesize
\setlength{\tabcolsep}{4.0pt}
\renewcommand{\arraystretch}{1.05}
\begin{tabular*}{\columnwidth}{@{\extracolsep{\fill}}l|cccc@{}}
\toprule
\textbf{Evidence} & \textbf{Overall} & \textbf{HLD} & \textbf{Calls/Q} & \textbf{Input tok./Q} \\
\midrule
Recent frames & 54.5 & 51.8 & 1.00 & 3.06K \\
+ direct memory & 52.5 & 75.0 & 1.00 & 3.54K \\
+ expanded memory & 55.5 & 71.4 & 1.72 & 4.28K \\
Full query policy & \textbf{56.7} & \textbf{80.4} & 2.37 & 6.29K \\
\bottomrule
\end{tabular*}
\end{table}

{\color{black}
\paragraph{Post-write entity matching.}
Table~\ref{tab:merge_main} shows that post-write entity matching primarily compresses and
organizes memory. Conservative matching reduces chains per video from 102.7 to
13.1, while accuracy changes from 54.5 to 53.0; more aggressive matching
reduces chains further without improving accuracy. A manual audit of 1,580
records in 155 chains found that 6.6\% of records were attached to the wrong
physical instance. Supplementary Table~\ref{tab:merge_policy_app} gives the
full policy and failure-type breakdowns.
}

\begin{table}[H]
\centering
\caption{\textcolor{black}{Post-write entity matching organizes and compresses
memory; it is not treated as an accuracy-improving module.}}
\label{tab:merge_main}
\color{black}
\footnotesize
\setlength{\tabcolsep}{5pt}
\renewcommand{\arraystretch}{1.0}
\begin{tabular*}{\columnwidth}{@{\extracolsep{\fill}}l|cc@{}}
\toprule
\textbf{Policy} & \textbf{Acc. (\%)} & \textbf{Chains/video} \\
\midrule
No entity matching & \textbf{54.5} & 102.7 \\
Conservative matching & 53.0 & 13.1 \\
Full matching & 53.0 & \textbf{11.9} \\
\bottomrule
\end{tabular*}
\end{table}

\subsection{\textcolor{black}{System Cost and Latency}}

{\color{black}
\paragraph{Streaming-memory maintenance.}
\method invokes one writer call per 8-second segment. With Qwen3-VL-8B, the
writer averages 5.88 seconds per segment on OVO-Bench and 7.68 seconds on
StreamingBench. Writing is asynchronous: queries use completed long-term
memory while the short-term visual buffer covers the most recent unfinished
segment. These are average-throughput measurements, not guarantees that every
segment finishes in real time. Full statistics and chunk-length sensitivity
appear in Appendix~\ref{app:system-efficiency}.

\paragraph{Representative question-time behavior.}
Direct retrieval is deterministic. VLM-assisted semantic query expansion is
invoked only when direct matching fails, while answer generation is invoked
once per query. On the 145-query StreamingBench diagnostic, \method averages
1.80 VLM calls per question. Call count alone does not
capture cost: the writer is multimodal, semantic expansion is a small text-only
catalog-linking call, and the answerer processes the final evidence prompt.
Table~\ref{tab:vlm_stages_main} summarizes these workloads.

\paragraph{ Controlled baseline comparison.}
Table~\ref{tab:matched_latency_main} reports \method and OASIS on the same three
streams and 50 questions using the same Qwen3-VL-8B model and hardware. This
small controlled run is used only for relative latency and token comparison;
the representative call frequencies are reported above. \method uses 72.2\%
fewer input tokens, yielding 4.05$\times$ lower TTFT and around 5 seconds lower
end-to-end latency. Supplementary Table~\ref{tab:vlm_call_breakdown_app} gives
the complete stage-level breakdown.
\textbf{End-to-end TTFT} is measured from query arrival through retrieval and
any semantic expansion to the first answer token; \textbf{answer-VLM TTFT}
measures only the answer call after its prompt is assembled.
}

\begin{table}[H]
\centering
\caption{\textcolor{black}{Question-time latency comparison with OASIS.}}
\label{tab:matched_latency_main}
\color{black}
\footnotesize
\setlength{\tabcolsep}{3.5pt}
\renewcommand{\arraystretch}{1.0}
\begin{tabular*}{\columnwidth}{@{\extracolsep{\fill}}l|ccc@{}}
\toprule
\textbf{Method} & \textbf{Input tok./Q} & \textbf{System TTFT} & \textbf{E2E} \\
\midrule
\method & 5,649 & 2.57s & 7.68s \\
OASIS & 20,331 & 10.41s & 12.19s \\
\bottomrule
\end{tabular*}
\end{table}

\begin{table}[H]
\centering
\caption{\textcolor{black}{VLM workloads in the streaming and query-time pipeline.}}
\label{tab:vlm_stages_main}
\color{black}
\footnotesize
\setlength{\tabcolsep}{3.2pt}
\renewcommand{\arraystretch}{1.0}
\begin{tabular*}{\columnwidth}{@{\extracolsep{\fill}}l|ccc@{}}
\toprule
\textbf{Stage} & \textbf{Invocation} & \textbf{Input tok.} & \textbf{Output tok.} \\
\midrule
Writer & 1/8s segment & 4,523/call & 781/call \\
Semantic expansion & Conditional & 681/trigger & 146/trigger \\
Answer generation & 1/query & 5,105/query & 351/query \\
\bottomrule
\end{tabular*}
\end{table}




\section{Conclusion}\label{sec:conclusion}
{\color{black}
We presented \method, a training-free focused semantic memory system for
streaming video QA. \method{} writes detailed records for selected entities and
actions while retaining compact context, then combines recent frames with
entity-centered long-term memory. Results on the evaluated OVO-Bench Perception
and Backward modes and on StreamingBench support focus-guided writing as an
effective way to allocate memory detail while controlling writing cost.
}


\section*{Acknowledgment}
This work is supported in part by the U.S. Army Research
Office (ARO) under grant W911NF-26-1-A164 and the U.S.
National Science Foundation (NSF) under grants
OAC-2505107 and OAC-2411446.
Distribution Statement A: \mbox{Approved} for public release. Distribution is unlimited.

{
    \small
    \bibliographystyle{ieeenat_fullname}
    \bibliography{ref}
}

\clearpage
\appendix
\section{Supplementary Overview}\label{sec:appendix}
This supplementary material provides dataset details, extended related work,
additional method details, extra quantitative results, qualitative analyses,
system-efficiency diagnostics, and prompt templates.
%
%
%
%
%
%



\section{Datasets}
\label{app:datasets}

This appendix describes the streaming-video benchmarks used in our main
experiments, with emphasis on the subsets over which we report results in
Section~\ref{sec:exp}.

\subsection{OVO-Bench}
\label{app:ovobench}

OVO-Bench~\cite{ovobench} is an online streaming-video question-answering
benchmark.
Given a text query $Q_{t_0}$ issued at time $t_0$ over a streaming video of
length $t_{\mathrm{end}}$, each question is restricted to a specific temporal
window of the stream, and the model must answer using only the evidence
available in that window.
Let $T$ be a recency threshold separating ``recent'' from ``earlier'' content.
OVO-Bench partitions its questions into three online video understanding
modes:
\textbf{Real-Time Visual Perception (RTV)}, which answers about the current
scene from the recent window $X_{[t_0 - T,\, t_0]}$;
\textbf{Backward Tracing (BT)}, which answers about earlier content from
$X_{[0,\, t_0 - T]}$; and
\textbf{Forward Active Responding (FAR)}, which decides when to respond as
future content $X_{(t_0,\, t_{\mathrm{end}}]}$ unfolds.
The full benchmark contains 2{,}814 QA pairs over 644 unique videos spanning
seven domains.
RTV and BT use a single-question multiple-choice protocol, while FAR uses a
different multi-query, time-aware decision protocol that triggers repeated
inference at densely sampled timestamps.
Following common practice among streaming peers, we evaluate on the full RTV
and BT subset of OVO-Bench -- the nine multiple-choice subtasks listed below,
totaling 1{,}468 questions across 512 unique videos -- and defer the three
FAR subtasks (REC, CRR, SSR) to future work.

\paragraph{Real-Time Visual Perception (RTV) subtasks.}
Real-Time Visual Perception evaluates whether the model can perceive,
comprehend, and reason about ongoing visual content at the current moment.
We report per-task accuracy on:
\begin{itemize}
    \item \textbf{OCR} -- Optical Character Recognition: recognize and
    interpret characters that appear within the frame (149 questions).
    \item \textbf{ACR} -- Action Recognition: recognize and interpret the
    actions being performed by individuals in the current frame (109).
    \item \textbf{ATR} -- Attribute Recognition: identify object
    characteristics such as color, texture, or size in nearby frames (116).
    \item \textbf{STU} -- Spatial Understanding: reason over spatial
    relationships between objects in nearby frames (178).
    \item \textbf{FPD} -- Future Prediction: forecast the most probable
    subsequent phase of the current scene, including changes in object
    states, actions, and other dynamic elements (101).
    \item \textbf{OJR} -- Object Recognition: recognize objects appearing in
    the current frames (184).
\end{itemize}

\paragraph{Backward Tracing (BT) subtasks.}
Backward Tracing evaluates whether the model can recall and reason about
events from earlier in the stream.
We report per-task accuracy on:
\begin{itemize}
    \item \textbf{EPM} -- Episodic Memory: backtrack and retrieve key
    moments from past video inputs (297 questions).
    \item \textbf{ASI} -- Action Sequence Identification: identify the
    correct ordering of human actions in the past stream (148).
    \item \textbf{HLD} -- Hallucination Detection: answer questions that are
    deliberately irrelevant to the existing video inputs, exposing
    confabulation behavior (186).
\end{itemize}

The RTV group probes whether the writer keeps the present scene addressable,
while the BT group probes whether the entity-centered memory preserves entity-
and event-level evidence beyond the recent visual window.
These two requirements directly correspond to the two roles of \method's
memory: focus-guided memory writing for current-moment salience, and the
persistent entity-centered memory for retrospective recall.

\subsection{StreamingBench}
\label{app:streamingbench}

StreamingBench~\cite{streamingbench} is a comprehensive streaming-video
understanding benchmark consisting of 900 videos and 4{,}500 human-curated
question-answer pairs across eight video categories.
Each video is paired with five questions presented at different timestamps,
and the model can only access the prefix observed before each question.
The benchmark organizes its 18 subtasks into three core aspects of streaming
video understanding:
\textbf{Real-Time Visual Understanding} (10 subtasks, 500 videos, 2{,}500
questions),
\textbf{Omni-Source Understanding} (4 subtasks, 200 videos, 1{,}000
questions), and
\textbf{Contextual Understanding} (4 subtasks, 200 videos, 800 questions).
Our table reports accuracy at the level the streaming literature most
commonly compares:
\begin{itemize}
    \item \textbf{Realtime} -- aggregate accuracy over the ten subtasks of
    Real-Time Visual Understanding: Object Perception, Causal Reasoning,
    Clips Summarization, Attribute Perception, Event Understanding,
    Text-Rich Understanding, Prospective Reasoning, Spatial Understanding,
    Action Perception, and Counting.
    These tasks evaluate whether the model can perceive, recognize, and
    reason about visual content as it appears in the current stream.
    \item \textbf{OmniSource} -- aggregate accuracy over the four subtasks
    of Omni-Source Understanding: Emotion Recognition, Scene Understanding,
    Source Discrimination, and Multimodal Alignment.
    These tasks require integrating synchronized visual and audio content
    within the stream.
    \item \textbf{SQA} -- Sequential Question Answering, a Contextual
    Understanding subtask in which each question is directly tied to an
    entity or event referenced by previous questions on the same video; the
    model must use episodic memory of the dialogue history to resolve
    cross-turn references.
    \item \textbf{Proactive} -- Proactive Output (PO), a Contextual
    Understanding subtask in which the model must autonomously decide
    \emph{when} to emit a prescribed output as the stream unfolds, rather
    than answering an explicit question.
    Evaluation uses a polling protocol that queries the model every second
    within a window around the ground-truth output time.
    \item \textbf{Overall} -- an aggregate score across the reported
    question categories.
\end{itemize}

Within each video sample, all timestamped queries are processed against a single
per-sample long-term semantic memory that is built incrementally as the stream
unfolds (Section~\ref{sec:method}).
Cross-turn entity binding therefore happens through the shared memory rather
than through prompt-only dialogue history, and previous-turn queries update
the focus state used when writing memory for later chunks
(multi-turn focus update).
We report Proactive Output for completeness, but interpret it separately from
the main explicit-query setting. Proactive Output primarily tests response timing,
whereas \method is designed to make previously observed entities and evidence
addressable for explicit queries.

\subsection{StreamBench}
\label{app:streambench}

StreamBench~\cite{streambench} is an open-ended streaming-video question
answering benchmark designed to evaluate multi-round interaction with memory
over time.
\textcolor{black}{We evaluate its 18-video, 107-question v0.3 subset; all
StreamBench results and conclusions below are scoped to this evaluated subset.}
Given a free-form natural-language question $Q_{t_0}$ posed at a specific
timestamp $t_0$ during a streaming video, the model must produce an
\emph{open-ended} answer using only the visual content observed up to $t_0$.
Unlike StreamingBench, no multiple-choice options are provided: predictions
are short natural-language sentences and are scored against ground-truth
references with an LLM-as-judge protocol.
The benchmark spans three video sources -- \textbf{Ego} (first-person
recordings), \textbf{WebVideo} (long-form web clips such as cooking shows and
outdoor tutorials), and \textbf{Movie} (short film segments) -- with a roughly
balanced number of videos per source.
Each video is paired with five or six breakpoint questions placed at
increasing timestamps, so that consecutive questions on the same video form a
natural multi-turn dialogue against an incrementally growing visual context.

StreamBench partitions its questions into six question subsets that probe
complementary memory and reasoning skills.
We report per-subset accuracy on:
\begin{itemize}
    \item \textbf{SF} -- Simple Factual: answer basic visual questions asked
    within the first 30 seconds of the video, testing short-term recall of
    readily observable facts.
    \item \textbf{OS} -- Object Search: locate a previously seen object on
    request (e.g.\ ``where can I find the black oven?''), requiring the model
    to recall a specific entity's last known position from memory rather than
    rediscover it in the current view.
    \item \textbf{SM} -- Short-term Memory Search: recall recently completed
    events, with less than 20 seconds between event completion and the query.
    \item \textbf{LM} -- Long-term Memory: recall facts or events from
    \emph{earlier} in the stream, often crossing scene boundaries or referring
    to objects no longer in view (e.g.\ ``where did I place the cutting board
    after washing it?'').
    \item \textbf{CI} -- Conversational Interaction: answer the current
    question using relevant dialogue from more than two turns earlier,
    testing memory of prior conversations.
    \item \textbf{KG} -- Knowledge Grounded: general-knowledge questions that
    are only loosely anchored to the video (e.g.\ ``what is the primary
    refrigerant used in refrigerators?'', ``why are vegetables green?''),
    answered primarily from world knowledge and largely independent of the
    visual stream.
\end{itemize}

Table~\ref{tab:streambench} reports per-subset accuracy alongside an unweighted
mean over the six displayed task scores.
The six subsets directly stress different facets of \method's memory pipeline:
SF and OS probe whether focus-guided memory writing keeps the current scene
addressable and whether the entity-centered memory retains a stable last-known
location per entity; SM and LM probe whether the memory preserves entity- and
event-level evidence across short and long temporal gaps, respectively; CI
probes whether earlier dialogue remains usable as context for later queries.

\section{Extended Related Work}
\label{app:extended-related-work}

This section provides the expanded related-work discussion corresponding to the
compressed overview in Section~\ref{sec:related}.

\subsection{Video-LLMs and Long-Video Understanding}

Large vision-language models have rapidly advanced from image understanding to
video understanding through stronger visual encoders, multimodal alignment, and
video instruction tuning~\cite{videollava,llavavideo,qwen2vl,qwen25vl,
qwen3vl,internvideo2,videollama,videochat}.
These models are commonly evaluated on offline video benchmarks that test
temporal reasoning, long-context understanding, and multimodal question
answering~\cite{mvbench,egoschema,videomme,mlvu,lvbench,nextqa}.
A complementary line of work improves long-video processing by scaling context
lengths or reducing visual tokens through sparse sampling, temporal pooling,
token compression, or adaptive frame selection~\cite{longva,llamavid,longvu,
videoxl,videoxltwo}.
Other systems introduce memory banks, hierarchical representations, or
retrieval-augmented reasoning over long videos~\cite{moviechat,malmm,
videostreaming,videoagent,videotree,goldfish,videorag}.

These works establish strong foundations for video-language reasoning, but they
largely operate in an offline regime: the model, compressor, or retriever has
access to the complete video, the query, or both before deciding which evidence
to use.
This assumption differs from streaming video understanding, where frames arrive
over time and future queries are unknown during memory construction.
\method therefore targets a different regime.
Rather than compressing a fully observed video for a known query, \method writes
memory online from the observed prefix and preserves the observed entities that
future queries may refer to.

\subsection{Streaming Video Understanding and Interaction}

Streaming video understanding has emerged to evaluate and build models that
process videos online and answer questions at arbitrary timestamps.
OVO-Bench studies real-time perception, backward tracing, and forward active
responding; StreamingBench evaluates real-time visual understanding,
omni-source understanding, and contextual interaction~\cite{ovobench,
streamingbench}.
SVBench, OVBench, RTV-Bench, \textcolor{black}{LiveStar}, and ESTP-Bench further broaden the
evaluation space toward temporal multi-turn dialogue, real-time perception,
online video dialogue, and just-in-time proactive response~\cite{svbench,
ovbench,rtvbench,livestar,eyeswideopen}.
These benchmarks show that streaming models must maintain temporal continuity
while remaining grounded in the current scene.

A growing set of Video-LLMs addresses this online setting through specialized
streaming architectures, training objectives, and interaction formats.
Early systems such as VideoLLM-online introduced streaming video dialogue with
response/silence decisions~\cite{videollmonline}, while subsequent models
improve stream processing through mixture-of-depth computation, online memory
buffers, frame-wise decoding, or streaming-aligned training~\cite{videollmmod,
streamchat,streamingvlm,streambridge,aura,livecc,
internlmomnilive}.
Another active direction focuses on proactive interaction: deciding when an
assistant should respond, remain silent, interrupt, or provide visual
instruction feedback~\cite{mmduet,livestar,eyeswideopen,lionfs,streammind,
vispeak,egospeak,proassist,omnimmi}.
Recent reasoning-oriented models also explore watching while thinking,
streaming chain-of-thought, and language traces as memory~\cite{
thinkwhilewatching2026,videostreamingthinking2026,
thinkingstreamingvideo2026,thinkasyousee2026,wat2026,
speakwhilewatching2026}.

This literature is relevant but studies a different axis from ours.
Many of these systems ask how to process streams efficiently, how to align
training with streaming inference, or when a model should speak.
\method instead asks what evidence should be written into memory so
that later questions remain answerable.
It is complementary to proactive-response systems: whereas they decide when to
respond, \method uses interaction to decide which entities should be remembered
in greater detail.

\subsection{Memory, Retrieval, and Compression for Streaming Video}

The closest line of work studies how to retain useful history under memory and
latency constraints.
A first family operates at the level of visual tokens, features, or KV caches.
QueryStream prunes tokens using query relevance and temporal novelty;
FluxMem, FreshMem, and CurveStream maintain adaptive hierarchical visual
memories using adjacency, frequency-space, or curvature-based criteria;
TimeChat-Online, STC, ReKV, StreamKV, InfiniPot-V, StreamMem, LiveVLM,
StreamingTOM, and related methods compress or retrieve visual tokens and KV
states to keep streaming inference efficient~\cite{querystream,fluxmem,
freshmem,curvestream,timechatonline,stc,rekv,streamkv,infinipotv,streammem,
livevlm,streamingtom,hermes}.
SimpleStream provides an important sanity baseline, showing that strong recent
frames alone can be competitive on streaming benchmarks~\cite{simplestream}.
Together, these works demonstrate that retention and compression are central to
streaming Video-LLMs.

A second family organizes history into temporal scenes, events, or hierarchical
memories.
OASIS maintains short and medium visual windows plus an on-demand hierarchical
event forest; StreamForest builds a persistent event-memory forest;
EventMemAgent adds adaptive tool use over event-centric memory; Vista
compresses and recalls scene units; Event-VStream and hierarchical
event-memory systems use event boundaries as the basic online unit~\cite{
oasis2026,streamforest,eventmemagent,vista,eventvstream,
hierarchicaleventmemory}.
Offline event-memory systems such as Video-EM similarly show the value of
structured episodic memories, although they build them with access to the full
video and query~\cite{videoem}.

\method differs from both families in the unit of memory.
Token and KV methods preserve computation states, but they do not expose stable
entities for later reference.
Scene and event memories preserve temporal units, but a single entity may
persist across multiple events, change names across descriptions, move between
locations, or be referenced later through an alias or answer option.
\method instead stores persistent entity identities and accumulates their
attributes, states, locations, relations, action chains, and evidence pointers.
This entity-level memory makes the stream addressable in the same terms used by
user questions.

\subsection{Structured Entity and World Memory}

Several recent methods move beyond flat visual tokens toward structured memory.
MA-LMM and MovieChat introduce memory banks for long-video understanding, while
VideoStreaming and VideoTree build compact or hierarchical representations for
query-driven reasoning~\cite{malmm,moviechat,videostreaming,videotree}.
VideoAgent, Goldfish, and Video-RAG-style systems treat long videos as
retrieval corpora, selecting relevant frames, clips, or auxiliary textual
signals at query time~\cite{videoagent,goldfish,videorag}.
More recent memory-agent approaches, such as WorldMM and MM-Mem, construct
multi-source memories, knowledge graphs, visual memories, or pyramidal semantic
memories for long-horizon video reasoning~\cite{worldmm,mmmem}.
These works show the promise of structured memory for video, especially when
questions require information distributed across time.

However, most structured-memory systems are offline, query-conditioned, or
agentic: they often build or traverse memory after the full video is available,
use heavier tool stacks, or rely on learned retrieval policies.
\method is designed for the online streaming regime.
It writes a long-term semantic memory as the stream unfolds, keeps
recoverable keyframe evidence for entity records, and selects evidence using
explicit targets, anchors, verbs, temporal scope, and answer-option
mentions.
Thus, \method brings structured memory into streaming video understanding at
the entity level: the central memory item is not a segment, token, scene, or
event, but an observed entity whose history remains nameable and retrievable.

\section{Additional Method Details}
\label{app:method-details}

This appendix provides algorithmic details for the method described in
Section~\ref{sec:method}.
The algorithms follow the notation used in the main text.
Experiment-specific constants, model choices, prompts, and decoding settings
are described in the experimental setup and prompt appendix.

\subsection{Online Long-Term Semantic Memory Construction}
\label{app:memory-algorithm}

Algorithm~\ref{alg:memory-construction} summarizes the construction of the
long-term semantic memory.
The short-term visual buffer is maintained separately at the frame level; this
algorithm details the long-term semantic memory, chain view, and
visual-evidence cache.
The algorithm processes the video prefix in time-ordered segments.
Each segment selects keyframes, writes structured records with the writer VLM,
merges records into the long-term memory, updates focus, and refreshes
the chain view.

\begin{algorithm}
\caption{Online long-term semantic memory construction}
\label{alg:memory-construction}
\begin{algorithmic}[1]
\REQUIRE prefix $V_{\leq t}$, writer model $\mathcal{W}_{\theta}$, segment length $\Delta$
\ENSURE $\mathcal{M}_t=(\mathcal{O}_t,\mathcal{G}_t,\mathcal{B}_t)$, focus $\mathcal{F}_t$
\STATE $\mathcal{O}_0,\mathcal{G}_0,\mathcal{B}_0,\mathcal{F}_0 \leftarrow \emptyset$
\STATE $\{C_i\}_{i=1}^{N} \leftarrow \textsc{Segment}(V_{\leq t},\Delta)$
\FOR{$i=1$ to $N$}
    \STATE $\mathcal{K}_i \leftarrow \textsc{SelectKF}(C_i,\mathcal{F}_{i-1})$
    \STATE $\Lambda_i \leftarrow \textsc{AssignWritingLevels}(\mathcal{F}_{i-1})$
    \STATE $\hat{\mathcal{U}}_i \leftarrow \mathcal{W}_{\theta}(\mathcal{K}_i,\Lambda_i)$
    \STATE $\mathcal{O}_i \leftarrow \textsc{Merge}(\mathcal{O}_{i-1},\hat{\mathcal{U}}_i)$
    \STATE $\mathcal{B}_i \leftarrow \textsc{UpdateCache}(\mathcal{B}_{i-1},\mathcal{K}_i,\hat{\mathcal{U}}_i)$
    \STATE $\mathcal{F}_i \leftarrow \textsc{UpdateFocus}(\mathcal{F}_{i-1},\mathcal{O}_i,\hat{\mathcal{U}}_i)$
    \STATE $\mathcal{G}_i \leftarrow \textsc{BuildChains}(\mathcal{O}_i)$
\ENDFOR
\STATE \textbf{return} $(\mathcal{O}_N,\mathcal{G}_N,\mathcal{B}_N),\mathcal{F}_N$
\end{algorithmic}
\end{algorithm}

\subsection{Keyframe Selection and Chain Views}
\label{app:keyframe-chain-details}

For each segment $C_i=[s_i,e_i]$, \method selects a compact keyframe set
$\mathcal{K}_i\subset C_i$.
The candidate set contains boundary frames, a middle frame, and a high-change
frame.
For probe frames $g_1,\ldots,g_m$ sampled inside the segment, visual change is
estimated by
\begin{equation}
    \delta_j = \tfrac{1}{HW}\sum_{h,w}|g_{j+1}(h,w)-g_j(h,w)|,
\end{equation}
and the high-change probe is selected by $j^\star_i=\argmax_j \delta_j$.
When a segment has low visual change and no focused target is at risk of
disappearing, the writer keeps only boundary frames; otherwise, it keeps the
full candidate set.

The merged entity memory is converted into a chain view
\begin{equation}
    \mathcal{G}_i =
    (\{\mathcal{O}_i(o)\}_o,
     \{\mathcal{C}^{\mathrm{act}}_i(o)\}_o,
     \mathcal{R}_i),
\end{equation}
where $\mathcal{O}_i(o)$ stores entity records with location periods,
$\mathcal{C}^{\mathrm{act}}_i(o)$ stores action chains, and $\mathcal{R}_i$
aggregates cross-entity relation triples.
These chain views are the default textual evidence source used by the
query-time retrieval module.

\subsection{Query-Time Hybrid Retrieval and Answering}
\label{app:routing-algorithm}

Algorithm~\ref{alg:routing} summarizes evidence selection.
The retrieval module parses the query and answer options, links query mentions
to entity identities, and assembles a compact evidence block.
When structured linking yields no candidate entity, the catalog linker maps
the query to entity identifiers from the existing catalog through semantic
query expansion.

\begin{algorithm}
\caption{Query-time evidence selection}
\label{alg:routing}
\begin{algorithmic}[1]
\REQUIRE $q=(u,\mathcal{Y}_q,t_q)$, memory $\mathcal{M}_{t_q}$, answer generator $\Psi_{\phi}$
\ENSURE answer $\hat{y}$
\STATE $z_q \leftarrow \textsc{Parse}(u,\mathcal{Y}_q)$
\STATE $\mathcal{O}_q \leftarrow \emptyset$
\FORALL{$o \in \mathcal{O}_{t_q}$}
    \STATE compute $R(o\mid q)$
    \IF{$R(o\mid q) > 0$}
        \STATE $\mathcal{O}_q \leftarrow \mathcal{O}_q \cup \{(o,R(o\mid q))\}$
    \ENDIF
\ENDFOR
\STATE sort $\mathcal{O}_q$ by score; keep top-$K$
\IF{$\mathcal{O}_q = \emptyset$}
    \STATE $\mathcal{O}_q \leftarrow \textsc{SemLink}(z_q,\textsc{Cat}(\mathcal{O}_{t_q}))$
\ENDIF
\STATE $\mathcal{E}_q \leftarrow \textsc{Assemble}(z_q,\mathcal{O}_q,\mathcal{G}_{t_q})$
\STATE select recent frames $\mathcal{K}^{\mathrm{rec}}_q$ and stored frames $\mathcal{K}^{\mathrm{cache}}_q\subseteq\mathcal{B}_{t_q}$ from the visual-evidence cache
\STATE $\hat{y} \leftarrow \Psi_{\phi}(q,\mathcal{E}_q,\mathcal{K}^{\mathrm{rec}}_q,\mathcal{K}^{\mathrm{cache}}_q)$
\STATE \textbf{return} $\hat{y}$
\end{algorithmic}
\end{algorithm}

\subsection{Entity Relevance Score}
\label{app:object-ranking}

The relevance score $R(o\mid q)$ computed in Algorithm~\ref{alg:routing} sums
weighted matches between the parsed query fields and the entity's identity
fields, minus a small penalty for generic background entities:
\begin{equation}
\begin{aligned}
    &R(o\mid q) =
    w_T \sum_{r\in\mathcal{T}_q} M(r,o)
    + w_H \sum_{h\in\mathcal{H}_q} M(h,o) \\
    &+
    w_Y \sum_{y\in\mathcal{Y}^{\mathrm{key}}_q} M(y,o)
    + w_V \sum_{v\in\mathcal{V}_q} M(v,\mathcal{C}^{\mathrm{act}}_{t_q}(o)) \\
    &- w_B\,B(o).
\end{aligned}
\end{equation}
Here $M(\cdot,\cdot)$ is a normalized matching function over canonical names,
aliases, categories, attributes, and head nouns, with synonym expansion for
common entity-name variants.
The term $M(v,\mathcal{C}^{\mathrm{act}}_{t_q}(o))$ checks whether the query
verb appears in the entity's action chain, and $B(o)$ is a small penalty for
irrelevant background entities.
Entities with positive scores are sorted and truncated to form the candidate set
$\mathcal{O}_q$.

\subsection{Entity Matching}
\label{app:object-matching}

The entity matcher is designed to preserve stable identities while allowing
natural language variation across segments.
For a new mention $x$ and an existing entity $o$, \method computes a score
using normalized names, aliases, head nouns, categories, and partial-string
matches:
\begin{equation}
\begin{aligned}
    &m(x,o) =
    \max\bigl\{
    \lambda_{\mathrm{name}} m_{\mathrm{name}},\;
    \lambda_{\mathrm{alias}} m_{\mathrm{alias}}, \\
    &\lambda_{\mathrm{head}} m_{\mathrm{head}},\;
    \lambda_{\mathrm{cat}} m_{\mathrm{cat}},\;
    \lambda_{\mathrm{sub}} m_{\mathrm{sub}}
    \bigr\}.
\end{aligned}
\end{equation}
Here $m_{\mathrm{name}}$ checks canonical-name equality,
$m_{\mathrm{alias}}$ checks alias equality, $m_{\mathrm{head}}$ checks head-noun
agreement, $m_{\mathrm{cat}}$ checks category agreement, and
$m_{\mathrm{sub}}$ checks containment between normalized names.
A match is accepted when $m(x,o)\geq \tau_m$.

The matcher also extracts a small set of discriminative modifiers
$D(x)$ and $D(o)$, such as colors or other identity-bearing attributes.
For weak matches based only on head nouns or partial names, the match is
suppressed when
\begin{equation}
    D(x)\cap D(o) = \emptyset, \quad D(x), D(o) \neq \emptyset.
\end{equation}
This prevents two distinct entities with the same head noun, such as
differently colored bottles, from being merged into one entity slot.

\subsection{Focus Update and Writing Levels}
\label{app:focus-details}

The focus update uses two groups of features.
The positive features include visibility, first appearance, reappearance,
persistence, event participation, movement, state change, manipulable-target
status, and interaction-conditioned relevance.
The negative features include absence, long disappearance, and static
background behavior.
The score update is
\begin{equation}
    p_i(o) =
    \clip_{[0,1]}(\gamma p_{i-1}(o) + \Delta_i(o)),
\end{equation}
where
\begin{equation}
    \Delta_i(o) =
    \boldsymbol{\alpha}^{\top}\boldsymbol{\phi}^{+}_i(o)
    - \boldsymbol{\beta}^{\top}\boldsymbol{\phi}^{-}_i(o).
\end{equation}
Here $\boldsymbol{\phi}^{+}_i(o)$ and $\boldsymbol{\phi}^{-}_i(o)$ are the
positive and negative feature vectors above. The coefficients are fixed method
hyperparameters rather than learned parameters.

Writing levels are assigned by a deterministic rule that maps each entity to
one of four labels used by the writer VLM:
\begin{equation}
    \lambda_i(o)=
    \begin{cases}
    \mathsf{focus},   & \text{high priority},\\
    \mathsf{support}, & \text{useful evidence},\\
    \mathsf{context}, & \text{scene anchor},\\
    \mathsf{drop},    & \text{otherwise}.
    \end{cases}
\end{equation}
Only a limited number of focus entities are listed for detailed writing, while
support and context entities are listed for compact writing.
The first-appearance evidence rule preserves open-world discovery for newly
visible manipulated entities, tools, containers, appliances, text regions, and
unusual entities.

\subsection{\textcolor{black}{Fixed Controller and Retrieval Configuration}}
\label{app:fixed-parameters}

{\color{black}
Tables~\ref{tab:focus-parameters-app} and
\ref{tab:retrieval-merge-parameters-app} report the fixed configuration used
throughout the evaluation. The values are shared across datasets; they are not
selected separately for individual benchmarks.
}

\begin{table}[!htbp]
\centering
\caption{\textcolor{black}{Focus-controller configuration. Additive updates are
applied after temporal decay and the final score is clipped to $[0,1]$.}}
\label{tab:focus-parameters-app}
\color{black}
\scriptsize
\setlength{\tabcolsep}{3pt}
\renewcommand{\arraystretch}{1.03}
\begin{tabularx}{\columnwidth}{@{}lXr@{}}
\toprule
\textbf{Group} & \textbf{Cue or parameter} & \textbf{Value} \\
\midrule
Carry-over & Previous Focus score ($\gamma$) & $\times0.75$ \\
\midrule
Positive & Visible in the current segment & $+0.10$ \\
Positive & First appearance & $+0.15$ \\
Positive & Reappearance after absence & $+0.20$ \\
Positive & Persistent for at least two segments & $+0.02$ \\
Positive & Persistent tool or container & $+0.08$ \\
Positive & Event participation or interaction & $+0.35$ \\
Positive & Movement or placement & $+0.25$ \\
Positive & State change & $+0.20$ \\
Positive & Tool or container prior & $+0.12$ \\
Multi-turn & Entity linked from a completed query & $+0.25$ \\
Multi-turn & Pending query term linked later & $+0.18$ \\
\midrule
Negative & Absent from the current segment & $-0.08$ \\
Negative & Absent for more than three segments & $-0.12$ \\
Negative & Static background without activity & $-0.12$ \\
\midrule
Labels & High / medium / low score & $0.65/0.35/0.15$ \\
Focus & Generic / interaction-bearing eligibility & $0.72/0.38$ \\
Support & Support or state-change cue / score & cue or $\geq 0.18$ \\
Context & Static anchor, persistence, or score & cue, $\geq2$, or $\geq0.08$ \\
\midrule
Writing & Detailed / compact target lists & $4/10$ \\
Writing & Object / event output limits per segment & $8/5$ \\
\bottomrule
\end{tabularx}
\end{table}

{\color{black}
The interaction-bearing Focus threshold applies only when the corresponding
binary cue is present. Completed-query links instead add the fixed score boosts
listed above; they do not change a threshold. Support and context eligibility
also accept the listed binary cues. The detailed and compact lists control
description length but do not close discovery: the writer may still report
newly visible manipulated objects and other salient entities, subject to the
per-segment output limit.
}

\begin{table}[!htbp]
\centering
\caption{\textcolor{black}{Fixed query-time retrieval and post-write entity-matching
configuration.}}
\label{tab:retrieval-merge-parameters-app}
\color{black}
\scriptsize
\setlength{\tabcolsep}{3pt}
\renewcommand{\arraystretch}{1.03}
\begin{tabularx}{\columnwidth}{@{}lXr@{}}
\toprule
\textbf{Module} & \textbf{Parameter or criterion} & \textbf{Value} \\
\midrule
Direct retrieval & Target / anchor / option / action ($w_T/w_H/w_Y/w_V$) & $10/5/3/2$ \\
Direct retrieval & Unmatched-background penalty ($w_B$) & $1$ \\
Direct retrieval & Maximum returned entities ($K$) & $5$ \\
Semantic expansion & Trigger & zero direct entities \\
Semantic expansion & Maximum returned existing identifiers & $5$ \\
Semantic expansion & Maximum retry count & $1$ \\
Cache recovery & Maximum entities / historical frames & $3/4$ \\
\midrule
Entity matching & Normalized canonical-name equality & $1.00$ \\
Entity matching & Stored-alias equality & $0.95$ \\
Entity matching & Head noun with equal category & $0.80$ \\
Entity matching & Head noun without category evidence & $0.60$ \\
Entity matching & Normalized-name containment & $0.50$ \\
Entity matching & Acceptance threshold ($\tau_m$) & $0.50$ \\
Entity matching & Conflicting explicit colors & reject \\
\bottomrule
\end{tabularx}
\end{table}

{\color{black}
Semantic expansion receives only the existing memory catalog and may return
only identifiers already present in that catalog. Entity matching first
normalizes punctuation, common modifiers, and a fixed alias dictionary. For
head-noun and containment matches, explicitly conflicting colors suppress the
match even when the numeric acceptance threshold is met.
}

\subsection{Multi-Turn Focus Update Details}
\label{app:multiturn-details}

This subsection expands the multi-turn focus mechanism described in
Section~\ref{sec:multiturn-extension}.
When query $q_j$ arrives, \method extracts a set of focus terms $\Gamma_j$
from its targets, anchors, and answer-option mentions.
Each term is matched against the current entity-centered memory.
Matched entities receive an interaction aspect in $\Omega_i(o)$ and a focus
boost
$p_i(o) \leftarrow \clip_{[0,1]}(p_i(o) + \eta_q\,\mu_j(o))$,
where
$\mu_j(o) = \mathbf{1}[o \in \mathrm{Match}(\Gamma_j, \mathcal{O}_i)]$
is the query-term match indicator.
Terms that do not yet match any entity are placed in a pending set
$\mathcal{P}^{\mathrm{pend}}_i$ and retried after later segments are written.
If a pending term later matches a newly observed entity, that entity receives a
delayed boost
$p_i(o) \leftarrow \clip_{[0,1]}(p_i(o) + \eta_p\,\nu_i(o))$
and the term is added as an alias when appropriate; here
$\nu_i(o) = \mathbf{1}[o \in \mathrm{Match}(\mathcal{P}^{\mathrm{pend}}_i, \mathcal{O}_i)]$
is the pending-term match indicator.

\subsection{Record Field Schemas}
\label{app:record-fields}

\paragraph{Persistent entity slot (long-term memory entry).}
Each slot in the long-term entity memory $\mathcal{O}_t$ stores
$o_j = (\mathrm{id}_j, n_j, \mathcal{N}_j, c_j, \mathbf{a}_j, \mathcal{X}_j,
\mathcal{E}_j)$: a stable identifier $\mathrm{id}_j$, canonical name $n_j$,
alias set $\mathcal{N}_j$, category $c_j$, attribute set $\mathbf{a}_j$,
observation sequence $\mathcal{X}_j$, and event sequence $\mathcal{E}_j$.
An observation records the time span, location, support or carrier entity
when visible, state, relations, textual evidence, confidence, and references
to entity-specific keyframes in $\mathcal{B}_t$.
An event records the time span, action type, summary, participants, and
changed entities.

The writer VLM emits three kinds of records per segment, which feed the
slot fields above via the merge step.

\paragraph{Detailed entity records} carry identity (canonical name, aliases,
category), attributes, location, support or carrier entity when visible,
state, relations, interactions, state changes, evidence text, and keyframe
references.
For text-bearing evidence (scoreboards, screen text, jersey numbers), the
writer additionally preserves the surface text as a cue.

\paragraph{Compact entity records} carry the same identity fields together
with a single short location/state/relation note and keyframe references.

\paragraph{Event records} summarize an action with its participants and any
changed entities, attached to the segment timestamp and the participating
entity identifiers.

\subsection{Query-Type Taxonomy}
\label{app:question-types}

The query-type field $\chi_q$ in the parsed query $z_q$ takes values from
a fixed taxonomy: spatial, historical-location, current-location, interaction,
attribute, yes/no, hallucination-detection, and concept queries.
Each type selects a different evidence-assembly policy
(Appendix~\ref{app:evidence-assembly}).

\subsection{Evidence Assembly by Query Type}
\label{app:evidence-assembly}

Evidence selection inside $\textsc{Assemble}$ is parameterized by the query
type $\chi_q$ from $z_q$.
The assembled block always includes the parsed query context, target and
answer-option support, and a brief scene overview; type-specific selection
adds:
\begin{itemize}
    \item \textbf{Interaction}: action-chain events and participant links
    for the targets and anchors.
    \item \textbf{Current-location}: the most recent location period of the
    target entity.
    \item \textbf{Historical-location}: the full location chain of the target
    entity up to $t_q$.
    \item \textbf{Attribute}: stored attributes and visual descriptors.
    \item \textbf{Spatial}: relation triples involving the target entity.
    \item \textbf{Yes/no and hallucination-detection}: presence evidence
    together with a coverage tag indicating whether the target appears in
    $\mathcal{O}_{t_q}$.
    \item \textbf{Concept}: indirect evidence obtained via catalog linking
    over the entity catalog.
\end{itemize}
Evidence lines can be tagged with answer-option support when an answer option
aligns with an entity or event in the assembled context.

\section{Additional Dataset Results}
\label{app:additional-results}

\textcolor{black}{StreamBench provides a complementary evaluation setting to
OVO-Bench and StreamingBench: its questions are open-ended, the task space is
more diverse, and the multi-turn interaction pattern is less constrained. We
evaluate the 18-video, 107-question v0.3 subset with an LLM-as-judge evaluator.
As shown in Table~\ref{tab:streambench}, \method{} obtains a 70.6 macro-average,
with its strongest subset results on Short-term Memory Search (SM), Long-term
Memory (LM), Simple Factual (SF), and Knowledge Grounded (KG). Published baselines
are quoted from their corresponding papers and use their reported frame
settings, so this table provides a complementary breadth diagnostic rather
than a matched head-to-head comparison.}

\begin{table*}[!htbp]
\centering
\caption{\textcolor{black}{Results on the evaluated 18-video, 107-question
StreamBench subset~\cite{streambench}.} We report accuracy (\%) on the six
question subsets. Baseline task scores are quoted from the corresponding
papers. \textbf{Avg} is recomputed as the unweighted mean of the six displayed
task scores for all methods.
Best score in each column is highlighted in \textbf{bold}.}
\label{tab:streambench}
\small
\setlength{\tabcolsep}{4pt}
\resizebox{\textwidth}{!}{%
\begin{tabular}{l|cccccc|c}
\toprule
\textbf{Model} &
\textbf{OS} & \textbf{LM} & \textbf{SM} & \textbf{CI} & \textbf{KG} & \textbf{SF} & \textbf{Avg} \\
\midrule
GPT-4o                                          & \textbf{60.5} & 61.2 & 64.4 & \textbf{72.3} & 93.9 & 74.7 & \textbf{71.2} \\
LLaMA-VID                                       & 33.9 & 38.2 & 44.1 & 58.4 & 76.9 & 57.1 & 51.4 \\
LLaVA-Hound                                     & 37.6 & 43.2 & 53.4 & 55.7 & 76.3 & 62.0 & 54.7 \\
LongVA                                          & 41.1 & 47.4 & 57.6 & 59.8 & 80.7 & 66.1 & 58.8 \\
MiniCPM-V2.6                                    & 37.6 & 51.9 & 43.7 & 65.7 & 66.2 & 64.2 & 54.9 \\
VILA-1.5                                        & 36.1 & 44.4 & 50.8 & 68.3 & 78.6 & 65.5 & 57.3 \\
InternVL2                                       & 38.5 & 46.6 & 50.9 & 67.6 & 81.0 & 62.2 & 57.8 \\
InternLM-XComposer2.5                           & 38.8 & 43.3 & 50.8 & 65.6 & 88.4 & 60.5 & 57.9 \\
MovieChat                                       & 18.6 & 20.4 & 26.5 & 42.3 & 67.2 & 35.8 & 35.1 \\
FreeVA                                          & 35.6 & 37.5 & 43.7 & 58.8 & 84.0 & 53.7 & 52.2 \\
Video-online                                    & 41.4 & 48.8 & 52.9 & 62.7 & 69.2 & 64.1 & 56.5 \\
Flash-VStream                                   & 37.1 & 44.5 & 48.6 & 58.1 & 66.4 & 59.2 & 52.3 \\
StreamChat                          & 40.7 & 43.6 & 47.1 & 63.0 & 89.1 & 73.8 & 59.6 \\
\midrule
Qwen2.5-VL-7B                         & 36.1 & 41.3 & 41.0 & 50.5 & 76.5 & 62.3 & 51.3 \\
\,+ OASIS~\cite{oasis2026}                          & 35.9 & 39.4 & 44.1 & 54.5 & 76.9 & 64.9 & 52.6 \\
\midrule
Qwen3-VL-8B                                      & 33.8 & 50.5 & 47.8 & 69.4 & 73.5 & 62.6 & 56.3 \\
\,+ OASIS~\cite{oasis2026}                          & 47.0 & 53.6 & 54.3 & 71.7 & 86.1 & 67.6 & 63.4 \\
\midrule
\textbf{Qwen3-VL-8B + \method}                  & 44.4 & \textbf{64.7} & \textbf{66.7} & 64.7 & \textbf{94.7} & \textbf{88.2} & 70.6 \\
\bottomrule
\end{tabular}
}\\[2pt]

\end{table*}

For Table~\ref{tab:streambench}, published baselines follow their reported
frame settings. \textcolor{black}{The quoted Qwen/OASIS rows use the 0.5\,fps
setting reported by OASIS, while \method uses 1\,fps online memory
construction.}

\noindent\textbf{Takeaway.}
\textcolor{black}{On the evaluated subset, \method{} obtains a 70.6
macro-average, compared with a recomputed 63.4 for OASIS on the same backbone.
Because the reported frame settings differ, we do not present the 7.2-point
difference as a matched improvement. The strong SF, SM, and LM results are
consistent with retaining current visual details, recent state changes, and
earlier evidence, respectively.}

\subsection{Additional Efficiency and Ablation Tables}
\label{app:additional-efficiency-ablation}

\begin{table}[!htbp]
\centering
\caption{Accuracy and memory-writing cost under fixed/adaptive keyframe
selection and uniform/focus-guided writing. This table mirrors the main-paper
comparison.}
\label{tab:memory_writing_cost_app}
\scriptsize
\setlength{\tabcolsep}{2pt}
\resizebox{\columnwidth}{!}{%
\begin{tabular}{lccccc}
\toprule
\textbf{Variant} & \textbf{Acc.} & \textbf{Input} & \textbf{Output} &
\textbf{Records/Q} & \textbf{Bank KB/Q} \\
 & \textbf{(\%)} & \textbf{tok./Q} & \textbf{tok./Q} & & \\
\midrule
Fixed + uniform & 62.7 & 201.5K & 38.4K & 259.7 & 226.8 \\
Adaptive + uniform & 63.2 & 126.7K & 38.3K & 259.6 & 226.5 \\
Fixed + focus-guided & 65.9 & 207.3K & 26.4K & 179.2 & 154.9 \\
\method & \textbf{67.1} & \textbf{134.7K} & \textbf{26.2K} & \textbf{178.3} & \textbf{153.4} \\
\bottomrule
\end{tabular}
}
\end{table}

\begin{table}[!htbp]
\centering
\caption{\textcolor{black}{Five-setting focus-threshold diagnostic on OVO-Bench
Backward Tracing. Only the detailed/compact writing thresholds change;
frames, writer, schema, and reader are fixed.}}
\label{tab:focus_sensitivity_app}
\color{black}
\scriptsize
\setlength{\tabcolsep}{3pt}
\begin{tabularx}{\columnwidth}{@{}X|ccc@{}}
\toprule
\textbf{Setting} & \textbf{Acc. (\%)} & \textbf{D/C per segment} & \textbf{KB/video} \\
\midrule
Threshold $\times0.5$ & 65.0 & 2.40/1.40 & 96.77 \\
Threshold $\times0.8$ & 66.7 & 2.43/1.48 & 98.62 \\
Default & 65.0 & 2.39/1.40 & 93.30 \\
Threshold $\times1.2$ & 65.0 & 2.37/1.50 & 99.52 \\
Threshold $\times1.5$ & 65.0 & 2.38/1.49 & 98.07 \\
\bottomrule
\end{tabularx}
\end{table}

\begin{table}[!htbp]
\centering
\caption{\textcolor{black}{Post-write entity-matching policies. Matching
substantially reduces chains per video but does not improve accuracy over
keeping segment-level records separate.}}
\label{tab:merge_policy_app}
\color{black}
\scriptsize
\setlength{\tabcolsep}{3pt}
\begin{tabularx}{\columnwidth}{@{}X|cc@{}}
\toprule
\textbf{Policy} & \textbf{Acc. (\%)} & \textbf{Chains/video} \\
\midrule
No entity matching & \textbf{54.5} & 102.7 \\
Exact-name matching & 53.5 & 14.3 \\
Exact-name + alias matching & 53.5 & 14.2 \\
Conservative matching & 53.0 & 13.1 \\
Full matching & 53.0 & 11.9 \\
Full matching without color check & 53.0 & \textbf{11.6} \\
\bottomrule
\end{tabularx}
\end{table}

{\color{black}
The manual merge audit covers 1,580 records in 155 chains from 30 videos.
Among them, 105 records (6.6\%) are attached to the wrong physical instance;
22\% of chains contain at least one such record, with 15.9\% mean contamination
within affected chains. Under-merging is approximately 1\%. The 105 incorrect
attachments comprise 52 lexical collisions, 38 repeated visually similar
instances, and 15 identity-drift cases.
}

\begin{table}[!htbp]
\centering
\caption{Historical component-level token-budget diagnostic retained for
completeness. Values use saved prompts and responses; image tokens are
approximated separately from prompt text and therefore should not be compared
directly with the matched latency measurements in
Table~\ref{tab:vlm_call_breakdown_app}.}
\label{tab:component_token_budget_app}
\scriptsize
\setlength{\tabcolsep}{3pt}
\begin{tabularx}{\columnwidth}{@{}X|cc@{}}
\toprule
\textbf{Component} & \textbf{Input tokens} & \textbf{Output tokens} \\
\midrule
Writer VLM & 6.4K & 572.4 \\
Semantic query expansion & 824.6 & 627.2 \\
Answerer & 1,071.0 & 147.2 \\
\bottomrule
\end{tabularx}
\end{table}

{\color{black}
Direct retrieval is deterministic and does not invoke a VLM.
VLM-assisted semantic query expansion is conditional.
The broader 145-query StreamingBench diagnostic averages 1.80 calls per
question. Table~\ref{tab:vlm_call_breakdown_app} reports the stage-level token
and latency statistics.
}

\begin{table}[!htbp]
\centering
\caption{\textcolor{black}{Stage-level VLM-call breakdown. Memory-writing
statistics are measured over StreamingBench, while semantic-expansion and
answer-generation statistics use the matched question-time cost comparison.
Expansion latency covers the complete pre-answer stage.}}
\label{tab:vlm_call_breakdown_app}
\color{black}
\scriptsize
\resizebox{\columnwidth}{!}{%
\begin{tabular}{l|cccc}
\toprule
\textbf{VLM stage} & \textbf{Frequency} & \textbf{Input tok.} & \textbf{Output tok.} & \textbf{Latency} \\
\midrule
Memory writing & 1/segment & 4,523/call & 781/call & 7.68s E2E \\
Semantic expansion &  Conditional & 681/trigger & 146/trigger & 2.97s pre-answer \\
Answer generation & 1/query & 5,105/query & 351/query & 0.12s TTFT \\
\bottomrule
\end{tabular}
}
\end{table}

Table~\ref{tab:memory_ablation_app} gives the numerical component-ablation
results shown in Figure~\ref{fig:memory_ablation_drop}. This ablation uses a
sampled subset of 145 evaluable StreamingBench queries rather than the full
benchmark. Main-benchmark accuracy is reported separately in
Table~\ref{tab:streamingbench_main}.

\begin{table}[!htbp]
\centering
\caption{Component ablation corresponding to
Figure~\ref{fig:memory_ablation_drop}. Drops are percentage-point decreases
from the full system.}
\label{tab:memory_ablation_app}
\footnotesize
\setlength{\tabcolsep}{4pt}
\begin{tabular*}{\columnwidth}{@{\extracolsep{\fill}}l|rr@{}}
\toprule
\textbf{Variant} & \textbf{Acc. (\%)} & \textbf{Drop (pp)} \\
\midrule
Full \method & \textbf{71.0} & -- \\
w/o Recent Window & 32.4 & 38.6 \\
w/o Focus State & 65.5 & 5.5 \\
w/o Structured Memory & 68.3 & 2.7 \\
w/o Adaptive Visual Writing & 68.3 & 2.7 \\
w/o Interaction Focus & 69.7 & 1.3 \\
\bottomrule
\end{tabular*}
\end{table}

\begin{table*}[!t]
\centering
\caption{Additional query-time evidence ablation on the type-proportional half
split of OVO-Bench Backward Tracing used in the original submission. S denotes
the short-term visual buffer, O the long-term semantic memory,
Exp denotes VLM-assisted semantic query expansion, and B denotes keyframe evidence from the
visual-evidence cache. Calls include
semantic query expansion when it is used. The \method row is computed over the
successfully completed queries from this evaluation split.
Input tokens, end-to-end latency, and TTFT are
measured over the full query-time pipeline.}
\label{tab:reader_ablation_app}
\small
\setlength{\tabcolsep}{6pt}
\resizebox{\textwidth}{!}{%
\begin{tabular}{lccccc}
\toprule
\textbf{Setting} & \textbf{Acc.} & \textbf{Calls/Q} &
\textbf{Input tok./Q} & \textbf{E2E/Q} & \textbf{TTFT/Q} \\
 & \textbf{(\%)} & & \textbf{(K)} & \textbf{(s)} & \textbf{(s)} \\
\midrule
S & 54.5 & 1.00 & 3.18 & 8.11 & 2.72 \\
O direct & 41.0 & 1.00 & 1.47 & 12.51 & 6.56 \\
O + Exp & 42.6 & 1.54 & 1.54 & 12.84 & 6.58 \\
B & 52.5 & 1.54 & 4.01 & 14.14 & 8.68 \\
S + B & 50.5 & 1.54 & 6.94 & 13.52 & 7.70 \\
S + O direct & 52.5 & 1.00 & 4.39 & 12.29 & 6.53 \\
S + O + Exp & 55.5 & 1.54 & 4.47 & 12.81 & 6.72 \\
S + O direct + B & 50.0 & 1.00 & 7.67 & 16.05 & 8.95 \\
S + O + Exp + B & 50.5 & 1.54 & 7.86 & 16.78 & 9.30 \\
\method & \textbf{56.7} & 1.54 & 4.84 & 12.89 & 6.80 \\
\bottomrule
\end{tabular}
}
\end{table*}

\section{Qualitative Case Studies}
\label{app:case-studies}

We include two representative examples to illustrate how the entity-centered memory
is used at query time, and where errors still arise. We include sample
identifiers to make the analysis traceable to the released metadata.

\paragraph{Case 1: OVO-Bench OCR, correct.}
(\texttt{video\_id=1442}, \texttt{question\_id=m14s100\_035}).
\emph{Query: Which brand name appears repeatedly on the side panels?}
GT = C / Yokohama, Pred = C / Yokohama.
The chunk-level memory explicitly records the visual text:
\begin{quote}\small
48--56s: \texttt{signboard}, attributes: \texttt{YOKOHAMA logo}, location:
along the track, evidence summary: \texttt{YOKOHAMA signboard is visible}.
\end{quote}
The merged entity-centered memory preserves this as a signboard entity whose later
state says that it ``displays YOKOHAMA branding.'' At query time, the
reader uses the query and answer options to rank relevant entities; semantic
query expansion maps the generic phrase ``brand name'' to the
signboard/text entity, and the answerer selects the option containing
Yokohama. This example shows the intended behavior: OCR evidence is first
written into chunk memory, then merged into a stable entity slot, and finally
retrieved by entity-level catalog linking.

\paragraph{Case 2: OVO-Bench HLD, failure.}
(\texttt{video\_id=375}, \texttt{question\_id=m14s100\_064}).
\emph{Query: Where was the bike seat before I opened it?}
GT = B / Unable to answer, Pred = D / on the scooter.
The observed stream shows motorcycle repair, and the memory contains related
but insufficient evidence:
\begin{quote}\small
0--8s: \texttt{motorcycle}, attributes include \texttt{black seat}; \quad
64--72s: person adjusts the motorcycle seat; \quad
128--136s: fuel tank is open.
\end{quote}
The query-time reader retrieves the motorcycle entry because semantic query
expansion links ``bike seat'' to the motorcycle/seat memory. This expansion is
useful when the surface wording differs from the memory name, but here it
also creates a failure mode: the retrieved memory proves that a motorcycle
seat exists, but it never establishes a prior location before it was opened.
The answerer over-infers from the retrieved entity and predicts ``on the
scooter'' instead of abstaining. This illustrates why hallucination-detection
queries require not only retrieving related entities, but also
checking whether the retrieved evidence directly supports the queried fact.

\noindent\textbf{Takeaway.}
These cases show that \method succeeds when the needed evidence is explicitly
written and retrieved, but fails when the memory is related yet
not evidentially sufficient.

\section{Failure Taxonomy Discussion}
\label{app:failure-taxonomy-discussion}

\begin{table}[!t]
\centering
\caption{Failure taxonomy on the final 8B runs. We review incorrect
predictions from OVO-Bench and StreamingBench. Percentages are computed within
each dataset's error set. Missing denotes absent memory evidence; Not retr.
denotes written evidence that was not retrieved; Insuff. and Not discrim.
denote evidentially insufficient and insufficiently discriminative retrieval,
respectively; Ref. denotes unresolved multi-turn references.}
\label{tab:failure-taxonomy}
\scriptsize
\setlength{\tabcolsep}{2pt}
\begin{tabularx}{\columnwidth}{@{}X|c|ccccc@{}}
\toprule
\textbf{Dataset} & \textbf{Errors} & \shortstack{\textbf{Miss-}\\\textbf{ing}} &
\shortstack{\textbf{Not}\\\textbf{retr.}} &
\shortstack{\textbf{Insuff.}} &
\shortstack{\textbf{Not}\\\textbf{discrim.}} &
\shortstack{\textbf{Ref.}} \\
\midrule
StreamingBench & 37 & 5 (13.5) & 0 (0.0) & 2 (5.4) & 29 (78.4) & 1 (2.7) \\
OVO-Bench & 25 & 9 (36.0) & 0 (0.0) & 1 (4.0) & 15 (60.0) & 0 (0.0) \\
\bottomrule
\end{tabularx}
\end{table}

Table~\ref{tab:failure-taxonomy} complements the qualitative cases with a
small error review of the final 8B runs. In the reviewed StreamingBench errors,
most errors are not empty-memory failures but cases where the system retrieves
plausible evidence that still does not sufficiently discriminate among the
answer choices, especially on emotion, misleading-context, and counting
questions. In the reviewed OVO-Bench errors, the failures are split between
missing evidence and insufficiently discriminative retrieval, especially on EPM
and ASI, where related entities are often retrieved but do not directly
establish the queried past state or action order. ``Written but not retrieved''
is rare in both datasets, suggesting that the current bottleneck is less often
finding a candidate memory item than deciding whether the retrieved evidence is
specific enough to support a unique answer. We note that the reviewed
StreamingBench errors include few SQA cases, so cross-turn reference failures
\textcolor{black}{are underrepresented relative to StreamingBench as a whole.}

The taxonomy in Table~\ref{tab:failure-taxonomy} suggests that the current
retrieval module is not the dominant bottleneck. In both datasets, pure
``written but not retrieved'' failures are rare. Instead, the remaining errors
mainly come from two sources. First, some needed evidence is never written into
memory at sufficient detail, especially for fine-grained states, past
locations, or action transitions. Second, the answerer often over-commits once
it sees retrieved evidence that is related but not sufficiently
discriminative to support a unique answer. The implication is that future
gains are more likely to come from improving memory fidelity and
evidence-sufficiency judgment than from only increasing retrieval breadth. In
particular, the StreamingBench results point to stronger calibration and
cross-turn evidence use, while the OVO results point more strongly to missing
fine-grained memory for backward questions such as EPM and ASI.

\section{System Efficiency and Memory Footprint}
\label{app:system-efficiency}

Table~\ref{tab:appendix-stage1-system-efficiency} reports system-level
statistics collected from the saved memory-construction logs. The writer VLM
processes one fixed 8-second video chunk at a time, corresponding to
about $60/8=7.5$ VLM calls per video minute; small deviations come from partial
boundary chunks and duration rounding. To make the storage cost easier to
read, Table~\ref{tab:appendix-stage1-system-efficiency} reports only the
segment-level text records and visual-evidence cache memory per video minute.
The text-memory number is the persisted JSON footprint of chunk-level records
before entity-level merging; the final merged bank used by retrieval is smaller
and is reported separately in Table~\ref{tab:memory_construction_cost_main}.

\begin{table*}[!htbp]
\centering
\caption{Memory-construction footprint on OVO-Bench and StreamingBench.
Segment records are persisted JSON chunk-level records before entity-level
merging; visual-evidence cache memory is measured from the stored keyframe
cache.}
\label{tab:appendix-stage1-system-efficiency}
\small
\setlength{\tabcolsep}{4.5pt}
\resizebox{\textwidth}{!}{%
\begin{tabular}{l|c|ccc|ccc}
\toprule
\textbf{Dataset} & \textbf{Model} &
\textbf{Video min} &
\textbf{Segment Records (KB/min)} & \textbf{Visual-Evidence Cache (KB/min)} &
\textbf{Writer latency} & \textbf{Input tok.} & \textbf{Output tok.} \\
\midrule
OVO-Bench & 8B & 428.5 & 151.1 & 3614.7 & 5.88s & 4409 & 734 \\
StreamingBench & 8B & 314.7 & 155.5 & 3645.4 & 7.68s & 4523 & 781 \\
OVO-Bench & 7B & 428.5 & 139.6 & 3614.7 & 12.15s & 5577 & 628 \\
StreamingBench & 7B & 314.7 & 150.4 & 3635.2 & 14.63s & 5719 & 700 \\
\bottomrule
\end{tabular}
}
\end{table*}

Table~\ref{tab:appendix-stage2-system-efficiency} reports query-time costs
supporting the main-paper system-cost analysis. These diagnostics use 100
OVO-Bench queries and 145 StreamingBench queries; component-ablation accuracy
is reported separately in Table~\ref{tab:memory_ablation_app}.
\emph{Semantic query expansion rate} denotes how often the
reader uses VLM-assisted semantic query expansion beyond direct lexical memory
lookup.
\emph{Cache trigger} denotes the fraction of queries for which linked
keyframes from the visual-evidence cache are added to the answering input.
TTFT is measured for the query-time reader.

\begin{table*}[!t]
\centering
\caption{Query-time retrieval and answering costs.
Semantic query expansion is the rate at which the reader uses VLM-assisted
semantic query expansion beyond direct lexical memory lookup, and cache trigger
is the rate at which linked keyframes from the visual-evidence cache are added
to the answering input.}
\label{tab:appendix-stage2-system-efficiency}
\small
\setlength{\tabcolsep}{4.5pt}
\resizebox{\textwidth}{!}{%
\begin{tabular}{l|c|cccc|ccc}
\toprule
\textbf{Dataset} & \textbf{Model} &
\textbf{Questions} & \textbf{VLM calls/Q} & \textbf{TTFT/Q} &
\textbf{TTFT/call} & \textbf{Semantic expansion} & \textbf{Memory hit} &
\textbf{Cache trigger} \\
\midrule
OVO-Bench & 8B & 100 & 2.15 & 6.58s & 3.06s & 40.0\% & 85.0\% & 0.0\% \\
StreamingBench & 8B & 145 & 1.80 & 7.31s & 4.06s & 75.2\% & 95.9\% & 2.1\% \\
OVO-Bench & 7B & 100 & 1.86 & 8.81s & 4.74s & 0.0\% & 85.0\% & -- \\
StreamingBench & 7B & 145 & 1.70 & 7.27s & 4.28s & 69.7\% & 96.6\% & -- \\
\bottomrule
\end{tabular}
}
\end{table*}

\paragraph{Chunk-length sensitivity.}
Figure~\ref{fig:efficiency_tradeoff} varies only the segment length used by the
memory writer on the StreamingBench evaluation subset. Shorter
segments write memory more frequently without improving accuracy over the
default 8-second setting. Longer segments reduce writer cost, but the 32-second
setting loses 6.0 points because each update must summarize a larger interval.

\begin{figure}[!htbp]
\centering
\scriptsize
\resizebox{0.98\columnwidth}{!}{%
\begin{tikzpicture}[x=0.88cm,y=0.32cm]
    \def\ybase{64}
    \def\ymax{73}
    \draw[gray!55] (0.2,\ybase) -- (6.2,\ybase);
    \draw[gray!55] (0.2,\ybase) -- (0.2,\ymax);
    \draw[gray!55] (6.2,\ybase) -- (6.2,\ymax);

    \foreach \y in {64,66,68,70,72} {
        \draw[gray!18] (0.2,\y) -- (6.2,\y);
        \node[anchor=east,gray!70] at (0.05,\y) {\y};
    }

    \foreach \y/\lab in {64/{0},67/{0.5},70/{1.0},73/{1.5}} {
        \draw[gray!55] (6.2,\y) -- (6.32,\y);
        \node[anchor=west,blue!55!black] at (6.42,\y) {\lab};
    }

    \foreach \x/\lab in {1.0/{2s},2.1/{4s},3.2/{8s},4.3/{16s},5.4/{32s}} {
        \node[anchor=north] at (\x,63.25) {\lab};
    }

    \fill[orange!80!black] (0.72,\ybase) rectangle (1.28,69.9);
    \fill[orange!80!black] (1.82,\ybase) rectangle (2.38,70.3);
    \fill[orange!80!black] (2.92,\ybase) rectangle (3.48,71.0);
    \fill[orange!80!black] (4.02,\ybase) rectangle (4.58,69.5);
    \fill[orange!80!black] (5.12,\ybase) rectangle (5.68,65.0);

    \node[anchor=south,orange!80!black] at (0.82,70.13) {69.9};
    \node[anchor=south,orange!80!black] at (2.1,70.53) {70.3};
    \node[anchor=south,orange!80!black] at (3.2,71.23) {\textbf{71.0}};
    \node[anchor=south,orange!80!black] at (4.3,69.73) {69.5};
    \node[anchor=south,orange!80!black] at (5.4,65.23) {65.0};

    \draw[blue!55!black,line width=0.85pt]
        (1.0,71.92) -- (2.1,68.20) -- (3.2,66.16) -- (4.3,65.08) -- (5.4,64.54);
    \foreach \x/\y in {
        1.0/71.92,
        2.1/68.20,
        3.2/66.16,
        4.3/65.08,
        5.4/64.54
    } {
        \node[blue!55!black] at (\x,\y) {\large $\blacklozenge$};
    }

    \node[anchor=north] at (3.2,62.05) {Segment length};
    \node[anchor=south,rotate=90,gray!70] at (-0.48,68.5) {Accuracy (\%)};
    \node[anchor=south,rotate=90,blue!55!black] at (7.42,68.5) {Writer tokens / video (M)};

    \fill[orange!80!black] (1.05,74.15) rectangle (1.28,74.55);
    \node[anchor=west] at (1.40,74.35) {Accuracy};
    \node[blue!55!black] at (3.05,74.35) {\large $\blacklozenge$};
    \node[anchor=west] at (3.25,74.35) {Writer tokens / video};
\end{tikzpicture}
}
\caption{Chunk-length sensitivity using Qwen3-VL-8B on StreamingBench. Bars
show accuracy, and blue markers show
memory-writer input tokens per video.}
\label{fig:efficiency_tradeoff}
\end{figure}

The main takeaway is that the text footprint remains manageable even before
entity-level merging. Across these runs, the persisted segment-level JSON
records are about 140--156 KB per video minute, and the final merged entity
bank is smaller because records referring to the same entity are consolidated.
Selected visual evidence is represented by timestamped keyframe references and
re-extracted from the source video when needed.
This footprint suggests that memory growth is not an immediate storage
bottleneck for hours-long online streams. At the measured rates, a 10-hour
online prefix corresponds to roughly 90 MB of segment-level text records. Even
if selected keyframes are materialized, the visual-evidence cache grows at
about 3.6 MB per video minute, or roughly 2.2 GB over a 10-hour prefix; when
cached frames are stored as timestamped references, the persistent footprint is
substantially smaller.

\subsection{Long-Horizon Online Memory Evaluation}
\label{app:late-prefix-diagnostic}

We further evaluate \method in a long-horizon online setting, where the memory
has been constructed over a long observed prefix before the query arrives. The
time length here refers to the online prefix accumulated before query arrival,
not the full video duration available in an offline setting. The
diagnostic examples are sampled from OVO-Bench cases with long query-time
prefixes: we first select examples whose observed prefix is at least 20
minutes, and include a small number of additional examples from the 15--20
minute range to cover more available long-prefix cases. The completed run
reported below contains 47 evaluated queries after Stage-2 filtering.

\begin{table}[!htbp]
\centering
\caption{Long-horizon online memory evaluation for \method. The run evaluates
FOLIO after long observed prefixes before query arrival and reports both
downstream accuracy and memory footprint. Writer token counts are omitted
because this diagnostic run did not log writer input/output tokens.}
\label{tab:late-prefix-diagnostic}
\scriptsize
\setlength{\tabcolsep}{2.5pt}
\begin{tabularx}{\columnwidth}{@{}X|ccccc@{}}
\toprule
\textbf{Variant} & \textbf{Acc.} & \textbf{Correct} &
\shortstack{\textbf{VLM}\\\textbf{calls/Q}} &
\shortstack{\textbf{Written}\\\textbf{rec./Q}} &
\shortstack{\textbf{Merged}\\\textbf{KB/Q}} \\
\midrule
\method & 61.7 & 29/47 & 1.55 & 624.5 & 585.2 \\
\bottomrule
\end{tabularx}
\end{table}

This diagnostic is not used as a primary benchmark result. Its purpose is to
check that the same memory-construction pipeline remains stable after the
online memory has accumulated over a long prefix. The run completed without
Stage-2 errors, reached a 100.0\% memory-hit rate, and used 1.55 VLM calls per
query on average. The merged memory remains below 0.6 MB per query on this
long-prefix subset, indicating that entity merging keeps the persistent memory
size manageable even when many segment-level records are written.

\FloatBarrier
\section{Additional Multi-Turn Analysis}
\label{app:sqa-diagnostic}

\begin{table}[!htbp]
\centering
\caption{SQA analysis for multi-turn focus-state updates on the complete
StreamingBench SQA split, containing 50 videos and 250 queries. Paper reader
uses the previous query history available at the current turn, while no query
history removes previous queries from the answering prompt. Runtime is measured
per query. Sem. exp. denotes semantic query expansion rate, and Cache fr./Q
denotes the average number of cache frames added per query.}
\label{tab:sqa_interaction_focus_app}
\scriptsize
\setlength{\tabcolsep}{2pt}
\begin{tabularx}{\columnwidth}{X|ccccc}
\toprule
\textbf{Setting} & \textbf{Acc.} & \textbf{Calls/Q} &
\textbf{Runtime/Q} & \textbf{Sem. exp.} & \textbf{Cache fr./Q} \\
\midrule
\method, paper reader & 54.8 & 1.87 & 9.89s & 87.2\% & 0.02 \\
\method, no query history & 54.0 & 1.87 & 9.79s & 87.2\% & 0.02 \\
w/o Interaction Focus, paper reader & 53.6 & 1.91 & 10.57s & 91.6\% & 0.16 \\
w/o Interaction Focus, no query history & 53.2 & 1.91 & 10.30s & 91.6\% & 0.12 \\
w/o Focus State, paper reader & 51.2 & 1.88 & 10.27s & 87.6\% & 0.11 \\
w/o Focus State, no query history & 51.2 & 1.88 & 10.27s & 87.6\% & 0.11 \\
\bottomrule
\end{tabularx}
\end{table}

Table~\ref{tab:sqa_interaction_focus_app} isolates the role of interaction
focus on the 50-video SQA subset, where later queries often refer back to
entities or actions from previous turns, such as asking what a previously
discussed person later did or where a referenced object moved. The paper reader
with query history performs best, but removing query history changes the full
system only slightly. In contrast, removing interaction focus drops accuracy
under both reader prompts. This suggests that the gain is not simply due to
exposing the answerer to previous queries; previous turns also help by guiding
later memory construction.

\FloatBarrier
\section{Demo Browser}
\label{app:demo-browser}

\begin{figure*}[!htbp]
    \centering
    \begin{subfigure}[t]{0.96\textwidth}
        \centering
        \includegraphics[width=\linewidth]{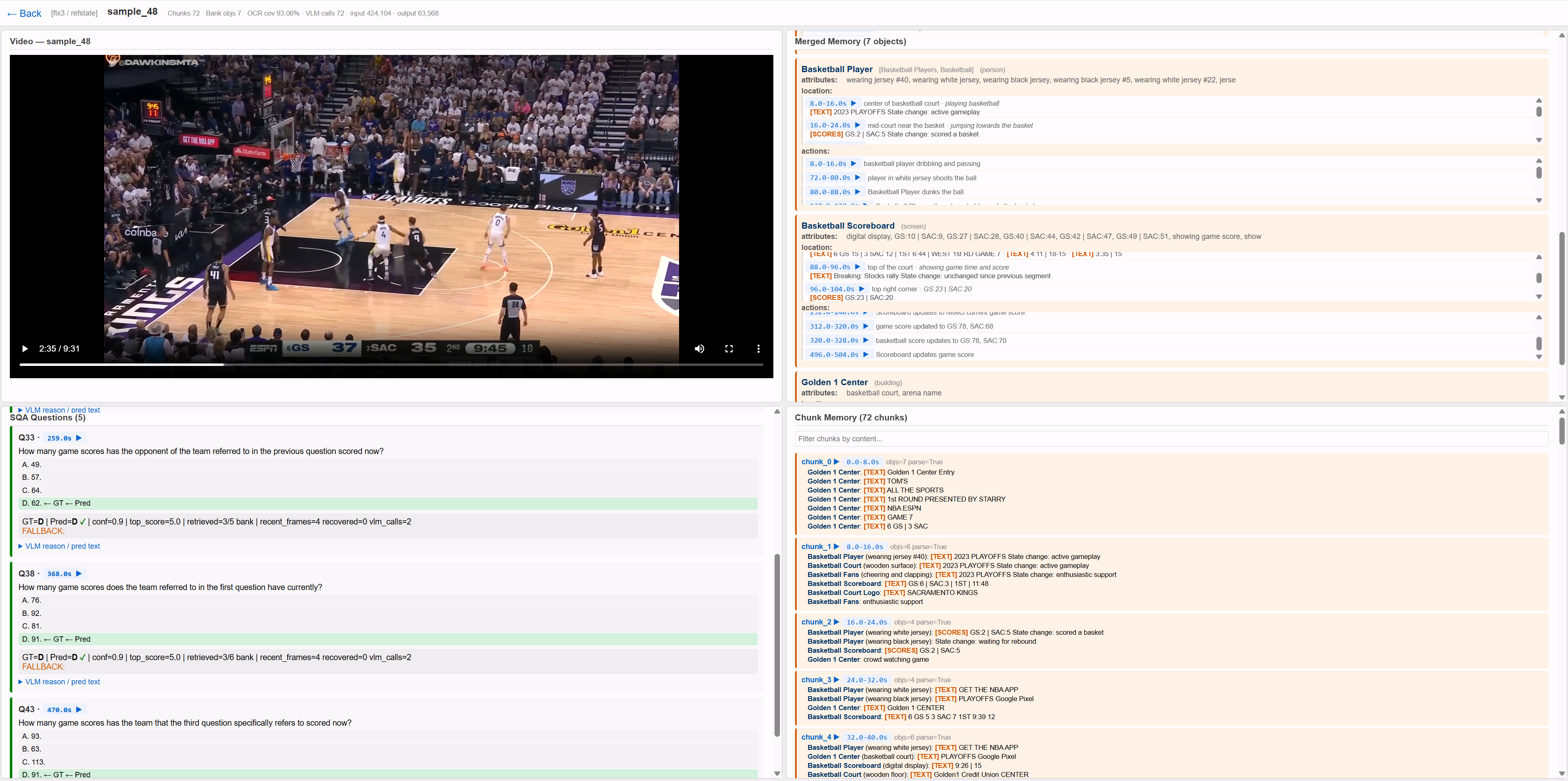}
        \caption{Basketball game example.}
        \label{fig:demo-browser-basketball}
    \end{subfigure}
    \vspace{0.6em}
    \begin{subfigure}[t]{0.96\textwidth}
        \centering
        \includegraphics[width=\linewidth]{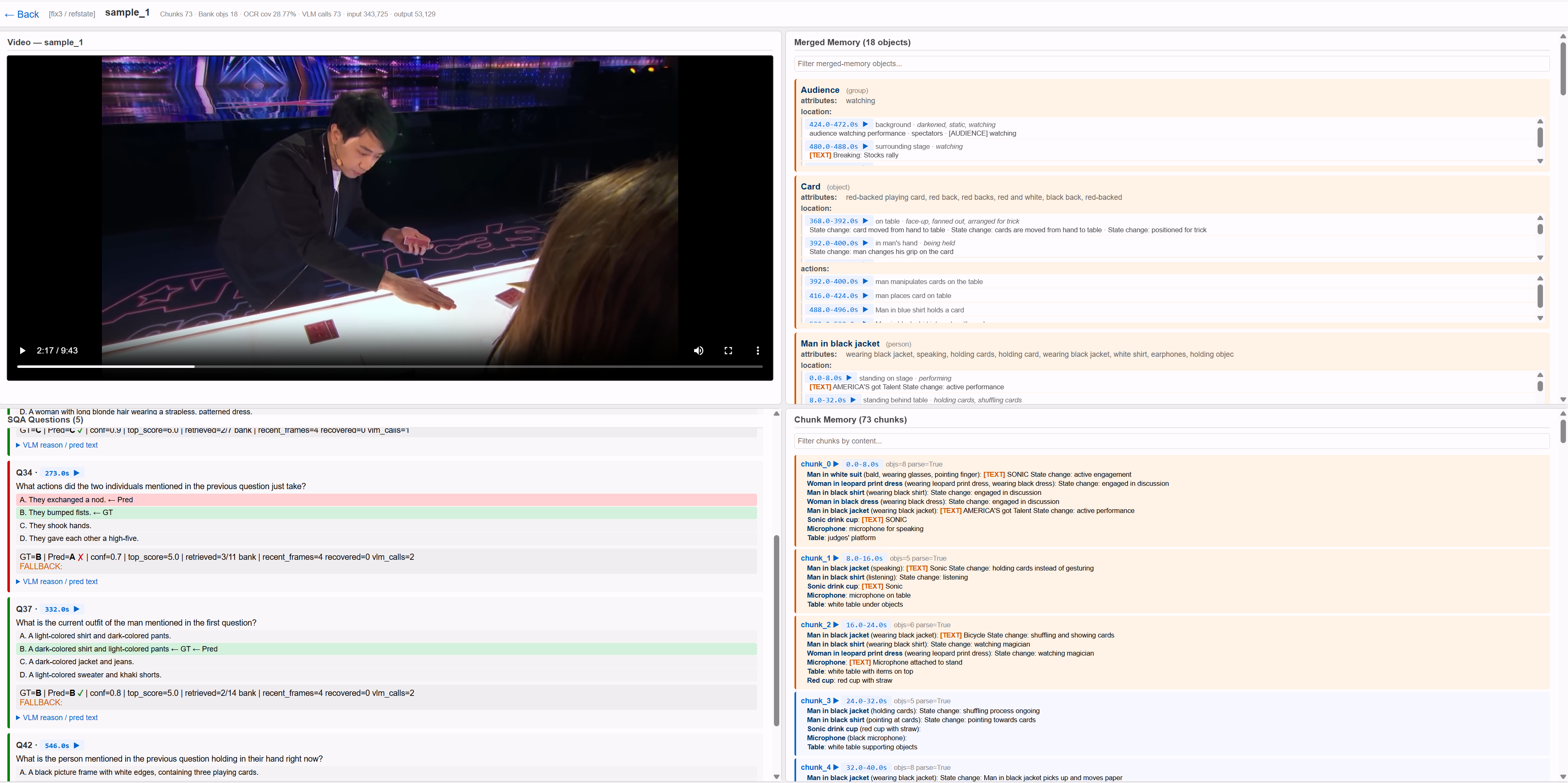}
        \caption{Variety and magic-show example.}
        \label{fig:demo-browser-magic}
    \end{subfigure}
    \caption{Interactive demo browser for inspecting video, queries, merged
    memory, and chunk-level memory in one view.}
    \label{fig:demo-browser}
\end{figure*}

Figure~\ref{fig:demo-browser} shows two screenshots of our internal browser
for inspecting \method on concrete streaming examples. The browser places the
video in the upper-left panel, the relevant query stream in the lower-left
panel, the merged entity memory in the upper-right panel, and the chunk-level
memory writes in the lower-right panel. We use this view to audit whether the
writer VLM preserved the correct entities, actions, text evidence, and
intermediate chunk responses, and whether the query-time reader retrieved
the intended support before producing an answer.

The same browser can be run in a streaming inspection mode. When the video is
scrubbed to time $t$, future questions, future chunks, and future merged-memory
updates are hidden, so the interface only exposes the questions that have
already arrived, the chunks already written, and the merged memory available by
time $t$. This makes it useful for checking that future information is hidden and diagnosing
whether an error comes from memory construction, retrieval, or answer
generation.

%

\clearpage
\onecolumn

\section{Prompt Templates}
\label{app:prompts}

The three vision-language calls used by \method are driven by the prompt
templates below, reproduced verbatim. Placeholders in braces (e.g.\
\texttt{\{start\_time:.1f\}}, \texttt{\{question\}}) are filled at call time
by the surrounding pipeline.

\subsection{Writer VLM Prompt}
\label{app:prompt-writer}

Used by the writer VLM $\mathcal{W}_{\theta}$ on every time-ordered segment: it
receives the selected keyframes and writing guidance with detailed and compact
object lists, and returns structured JSON containing detailed object records,
compact object records, and events.

\begin{promptbox}{Writer VLM Prompt (verbatim)}
Analyze this segment for budgeted object-state memory.

Time range: {start_time:.1f}s - {end_time:.1f}s

{object_state_budget}

The budget controls writing detail, not object discovery.

Required behavior:
1. DETAILED objects: if visible, write structured state with location, holder, state,
   relation, interaction, state_change, evidence_summary, and evidence_frames.
2. COMPACT objects: if visible, write one compact note with location/state/relation
   and evidence_frames.
3. Always discover newly visible manipulated objects, tools, containers, appliances,
   cooking surfaces, sinks, screens/text, and unusual objects. Put them in compact_objects
   unless they are central to the action, then put them in detailed_objects.
4. Do not ignore new or rare objects just because they are not listed in the budget.
5. Do not duplicate the same physical object across detailed_objects and compact_objects.

Return ONLY valid JSON:
{
  "time": "{start_time:.1f}-{end_time:.1f}s",
  "detailed_objects": [
    {
      "name": "object name",
      "category": "category",
      "attributes": ["attr1"],
      "location": "where it is",
      "holder": "none or holder",
      "state": "current visible state",
      "relations": [
        {"relation": "on/in/next_to/used_with/etc", "target": "other object"}
      ],
      "interactions": [
        {"type": "held/moved/used/etc", "with": "person/tool", "summary": "description"}
      ],
      "state_change": "what changed since previous segment, or stable",
      "evidence_summary": "specific visible evidence",
      "evidence_frames": [0],
      "confidence": 0.85
    }
  ],
  "compact_objects": [
    {
      "name": "non-focus visible object",
      "category": "category",
      "location": "compact location",
      "state": "compact state",
      "relation": "relation to action or detailed object",
      "brief": "one short useful memory note",
      "evidence_frames": [0],
      "confidence": 0.7
    }
  ],
  "events": [
    {
      "event_type": "action type",
      "summary": "what happened",
      "participants": ["obj1", "obj2"],
      "changed_objects": ["obj1"],
      "confidence": 0.8
    }
  ]
}

Rules:
1. Only list visible or partially visible objects.
2. Maximum {top_k_objects} total objects across both object lists and {top_m_events} events.
3. Prefer detailed budget for repeated or actively manipulated focus objects.
4. Preserve compact notes for tools, containers, appliances, cooking surfaces, sinks,
   screens/text, and newly appearing objects even when they are not the action center.
5. evidence_frames can be approximate local frame indices inside this segment; use an
   empty list if you cannot localize the evidence.
\end{promptbox}

\subsection{Semantic Query Expansion Prompt}
\label{app:prompt-semlink}

Invoked only when direct structured-memory retrieval returns an empty
candidate set, typically on abstract or concept-level queries. The VLM receives
the query, answer candidates, a compact object catalog, and an action catalog,
and is constrained to return existing object identifiers plus a short rationale
and a candidate-answer prior.

\begin{promptbox}{Semantic Query Expansion Prompt (verbatim)}
You are a video memory assistant. The rule-based retriever could not find direct keyword matches for this question. Use SEMANTIC REASONING to identify conceptually relevant objects.

## QUESTION
{question}

## OPTIONS
{options_text}

## ALL TRACKED OBJECTS IN MEMORY
{object_list}

## NOTABLE ACTIONS / EVENTS
{action_list}

## TASK
The question may use abstract concepts (e.g., "avid reader", "active sport") that don't directly name objects.

Your job:
1. Identify which tracked objects/actions are CONCEPTUALLY related to the question.
   - Example: "avid reader" -> bookshelf, books, magazine, reading lamp
   - Example: "musician" -> piano, guitar, drum
   - Example: "what activity is happening" -> look at action list
2. Return their object_ids (3-7 most relevant).
3. Also explain briefly which option seems most supported.

## OUTPUT (JSON only, no other text)
{
  "relevant_object_ids": ["obj_0001", "obj_0002", "obj_0005"],
  "reasoning": "Brief explanation of why these objects relate to the question concept.",
  "suggested_option": "A/B/C/D"
}
\end{promptbox}

\subsection{Answer Prompt}
\label{app:prompt-answerer}

Used by the answering VLM $\mathcal{A}_{\phi}$ at query time. It receives
the query, answer candidates, the consolidated memory block, recent
frames, and any retrieved cache frames; it returns a structured response
with the predicted label, supporting evidence, reason, and confidence. The
prompt's ``Mode B'' branch is auto-enabled when the consolidated memory
carries a concept-query header (set when semantic query expansion is used at
retrieval time).

\begin{promptbox}{Answer Prompt (verbatim)}
You are a video QA assistant. Answer the multiple-choice question using:
1. The RECENT FRAMES shown above (current visual state near question time)
2. The CONSOLIDATED MEMORY below (chronological history from earlier in the video)

The memory is built from object observations across the video, organized as time-ordered location chains and action chains.

MODE SELECTION (auto-detected from memory):

=== MODE A -- STRICT FACTUAL (default) ===
Use this mode UNLESS the memory contains the header "## (!) CONCEPT QUESTION -- SPECIAL HANDLING REQUIRED".

Rules:
1. Memory is the PRIMARY source for past events. Recent frames inform CURRENT state only.
2. For multiple choice, select one option label (A, B, C, or D).
3. If "Unable to answer", "cannot determine", "not visible", or similar is among the options AND memory/frames do NOT contain DIRECT, EXPLICIT evidence, you MUST select it.
4. Avoid speculation: prefer "Unable to answer" (when available) over guessing.
5. BANNED words: "most likely", "suggests", "implies", "could be", "commonly", "typically", "probably". If used, answer is likely "Unable to answer".
6. Mere presence of an object does NOT confirm its location, action, or relation.

=== MODE B -- CONCEPT REASONING (only when memory has CONCEPT QUESTION header) ===
This mode is enabled by the rule-based retriever failing entirely on this question (suggesting it asks an abstract concept like "is a reader" / "is a musician").

Rules:
1. INDIRECT evidence IS valid: bookshelf -> reader; piano -> musician; tools -> handyman.
2. You do NOT need direct visual confirmation of the activity.
3. The BANNED-words rule of Mode A does NOT apply here -- you may use "suggests/implies".
4. Trust the LLM-suggested option (if any) unless contradicted.

=== COMMON ===
Trust the QUESTION CONTEXT and OPTION CHECK signals at the top of memory when present.
Return ONLY valid JSON.

## QUESTION
{question}

## OPTIONS
{options_text}

## CONSOLIDATED MEMORY
{memory_text}

## DECISION PROCESS
- Step 1: Does memory/frames EXPLICITLY state the answer? If yes -> pick that option.
- Step 2: If memory says "TARGET NOT FOUND" -> pick "Unable to answer" if available.
- Step 3: Else, pick the option least contradicted by evidence.

## OUTPUT JSON FORMAT
{
  "prediction_label": "A/B/C/D",
  "prediction_text": "the text of your chosen option",
  "visual_observation": "what you see in recent frames",
  "supporting_evidence": [
    {"source": "memory/frames", "time": "...", "evidence": "..."}
  ],
  "reason": "brief explanation",
  "confidence": 0.0-1.0
}
\end{promptbox}

\end{document}